\documentclass[a4paper,11pt]{article}
\usepackage[margin=1in]{geometry} 

\usepackage[utf8]{inputenc}
\usepackage[T1]{fontenc}

\usepackage{cite}
\usepackage{amsmath,amssymb,amsfonts}
\usepackage{graphicx}
\usepackage{textcomp}
\usepackage{xcolor}
\usepackage{hyperref} 
\usepackage{url} 
\usepackage{booktabs}
\usepackage{multirow}
\usepackage{xspace}
\usepackage{algorithm}
\usepackage{algpseudocode}
\usepackage{tabularx}
\usepackage{amsmath} 
\usepackage{enumitem}

\usepackage{tikz}
\usepackage{collcell}
\usepackage{calc}

\newcommand*{\MinNumber}{0.1}%
\newcommand*{\MaxNumber}{0.5}%

\newcommand{\ApplyGradient}[1]{%
    \pgfmathsetmacro{\PercentColor}{max(min(100.0*(#1 - \MinNumber)/(\MaxNumber-\MinNumber) * 0.5,100.0),0.00)}\hspace{-0.13em}\colorbox{gray!\PercentColor!white}{#1}
}

\newcolumntype{R}{>{\collectcell\ApplyGradient}c<{\endcollectcell}}

\newcommand{\beginsupplement}{%
    \renewcommand{\thefigure}{S\arabic{figure}}
    \renewcommand{\thetable}{S\arabic{table}}
    \renewcommand{\thealgorithm}{S\arabic{algorithm}}
    \renewcommand{\thesection}{\Alph{section}} 
    \setcounter{figure}{0}
    \setcounter{table}{0}
    \setcounter{algorithm}{0}
    \setcounter{equation}{0}
    \setcounter{section}{0}
}

\newcommand{\hide}[1]{}

\newcommand{\eg}{{\textit{e.g.}}}

\newcommand{\model}{\textsc{DyGETViz}\xspace}

\newcommand{\problem}{dynamic graph embedding trajectories visualization\xspace}

\title{Empowering Interdisciplinary Insights with Dynamic Graph Embedding Trajectories}

\author{
    Yiqiao Jin$^{1}$,
    Andrew Zhao$^{1}$,
    Yeon-Chang Lee$^{2}$, \\
    Meng Ye$^{3}$, 
    Ajay Divakaran$^{3}$, 
    Srijan Kumar$^{1}$ \\
    $^{1}$Georgia Institute of Technology, \\
    $^{2}$Ulsan National Institute of Science and Technology (UNIST), \\
    $^{3}$SRI International \\
    \texttt{\{yjin328,srijan\}@gatech.edu} \\
}

\date{}

\begin{document}

\maketitle

\begin{abstract}
We developed \model, a novel framework for effectively visualizing dynamic graphs (DGs) that are ubiquitous across diverse real-world systems. 
This framework leverages recent advancements in discrete-time dynamic graph (DTDG) models to adeptly handle the temporal dynamics inherent in dynamic graphs. \model effectively captures both micro- and macro-level structural shifts within these graphs, offering a robust method for representing complex and massive dynamic graphs. 
The application of \model extends to a diverse array of domains, including ethology, epidemiology, finance, genetics, linguistics, communication studies,
social studies, and international relations. 
Through its implementation, \model has revealed or confirmed various critical insights. These include the diversity of content sharing patterns and the degree of specialization within online communities, the chronological evolution of lexicons across decades, and the distinct trajectories exhibited by aging-related and non-related genes. Importantly, \model enhances the accessibility of scientific findings to non-domain experts by simplifying the complexities of dynamic graphs. 
Our framework is released as an open-source Python package for use across diverse disciplines. Our work not only addresses the ongoing challenges in visualizing and analyzing DTDG models but also establishes a foundational framework for future investigations into dynamic graph representation and analysis across various disciplines.
\end{abstract}

\flushbottom

\section{Introduction}
\label{sec:introduction}



\vspace{1mm}
\paragraph{Background} 
Dynamic graphs (DGs) are ubiquitous data structures present in various real-world evolving systems, such as social networks~\cite{jin2023predicting}, linguistics~\cite{hamilton2016diachronic}, international relations~\cite{monken2021graph}, and computational finance~\cite{huang2022dgraph}. 
Representing these dynamic graphs efficiently has become a crucial challenge due to their massive sizes and ever-changing nature. 
One compelling approach to tackle this challenge is \emph{discrete-time dynamic graph} (DTDG) models~\cite{you2022roland,pareja2020evolvegcn, seo2018structured}, which represent a dynamic graph as a series of snapshots, each containing the nodes and edges that co-occur at particular timestamps. 
Despite the effectiveness of DTDG models in a wide range of graph-oriented tasks such as link prediction, node classification, and edge regression, these models usually remain opaque to researchers in terms of interpretability. 
The high-dimensional representations generated by these models make it difficult for users to extract and understand the intrinsic value from dynamic graphs. 
Currently, researchers often manually analyze the dynamic graph data, as there are no specialized tools to support this process~\cite{li2020argo,mersch2013tracking}. However, manual analysis of enormous dynamic graphs covering multiple timestamps can be overwhelming, and the continuously evolving nature of these graphs makes it challenging to intuitively capture both \emph{micro-level} and \emph{macro-level} structural shifts. 
For instance, in the study of international relations, aside from predicting graph attributes like future bilateral trade volumes, it is vital to understand \emph{micro-level} changes such as a country's alliance network, trade relations, and conflict dynamics, as well as \emph{macro-level} trends such as the stability of the global economy amidst wars and financial crises, as inherently reflected by the high-dimensional node embeddings obtained from DTDG models. 

In this case, \emph{visualization} becomes a powerful tool with an intuitive and user-friendly interface for analyzing the dynamic graph embeddings of DTDG models, as it enables researchers to gain both \emph{micro-level} understandings, such as predicting node states and future trajectories, and \emph{macro-level} analysis, such as forecasting emerging turning points in geopolitical events. 
With an effective visualization framework, researchers can gain insights, identify patterns, detect anomalies, and effectively communicate their findings to both domain experts and the general public, which would be challenging to achieve solely through manual analysis. 




\vspace{1mm}
\paragraph{Challenges}
Developing a visualization framework for dynamic graph embedding trajectories
requires addressing the unique characteristics and challenges of DGs. 
The first challenge is the constant addition and deletion of nodes in DTDG. 
As nodes are continuously added or removed, accurately inferring dynamic embedding trajectories for new nodes and effectively incorporating them into the visualization becomes crucial yet challenging.
More specifically, the continuous addition and removal of nodes create a dynamic landscape in which the proximity between nodes is in constant flux, further complicating the visualization process. 
The second challenge arises from the persistent evolution of node embeddings over time. 
Conventional visualization techniques~\cite{hotelling1933analysis,van2008visualizing,pezzotti2016hierarchical} often rely on non-parametric methods, which can present limitations when projecting new data points onto an existing visualization space~\cite{an2020viva}. When applying such visualization techniques to each snapshot of the DTDG, the visualization layout undergoes a complete transformation, disrupting the continuity and hindering a coherent representation of embedding trajectories over time~\cite{an2020viva}. Thus, researchers will fail to observe valuable patterns in the DTDG network. Addressing this challenge is crucial for providing researchers with a clear and consistent understanding of the dynamic graph's behavior and evolution over time. 

\vspace{1mm}
\paragraph{This Work}
In this work, we formally define the novel problem of \problem to enable the analysis of discrete-time dynamic graph models. We propose \model, a novel framework for \underline{Dy}namic \underline{G}raph \underline{E}mbedding \underline{T}rajectory  \underline{Vi}suali\underline{z}ation, to address the above challenges. \model leverages recent developments in dynamic graph neural networks (GNNs)~\cite{seo2018structured,chen2022gc,sankar2020dysat} and offers two key functionalities: \textit{visualization} and \textit{analytics}. The visualization module employs principles from dynamic GNNs to map high-dimensional node embeddings into lower-dimensional representations, and employs a flexible and computationally efficient approach to project node state at each timestamp onto the visualization, which is potentially scalable to datasets spanning multiple timestamps. 
The analytics module quantifies structural shifts in DTDGs from both micro- and macro-level. For micro-level analysis, it uses two similarity measures, namely, Jaccard index~\cite{jaccard1912distribution} and Rank-biased Overlap (RBO)~\cite{webber2010similarity,oh2022rank}, to quantify the changes in the local topology of each node between adjacent timestamps. For macro-level analysis, it uses a novel metric, normalized average ranking change (NARC), as well as the absolute volumes of embedding movements to assess the changes in global topology. These comprehensive analytics enable researchers to gain insights into both fine-grained and large-scale changes in dynamic graphs, empowering investigations across various domains. 
The versatility and applicability of \model is demonstrated by our analysis on nine datasets introduced in Supp.\ref{ref:dataset} spanning different graph sizes and domains, including ethology, epidemiology, finance, genetics, linguistics, communication studies, social studies, and international relations.

We provide complete technical details for \model in Supp.\ref{sec:method}. 
Our proposed python package is available at \href{https://github.com/claws-lab/dygetviz.git}{\underline{GitHub}}, and the visualization for all datasets are available on \href{http://dygetviz.eastus.cloudapp.azure.com/}{\underline{our website}}.  
All the code and datasets have been made publicly available.

\section{Results}\label{sec:experiments}
\subsection{Reddit Community Graphs Reveals Content Specialization, Content Diversity, and Echo Chambers } 
\label{sec:Reddit}
\begin{figure*}[ht]
    \centering
    \includegraphics[width=0.95\textwidth]{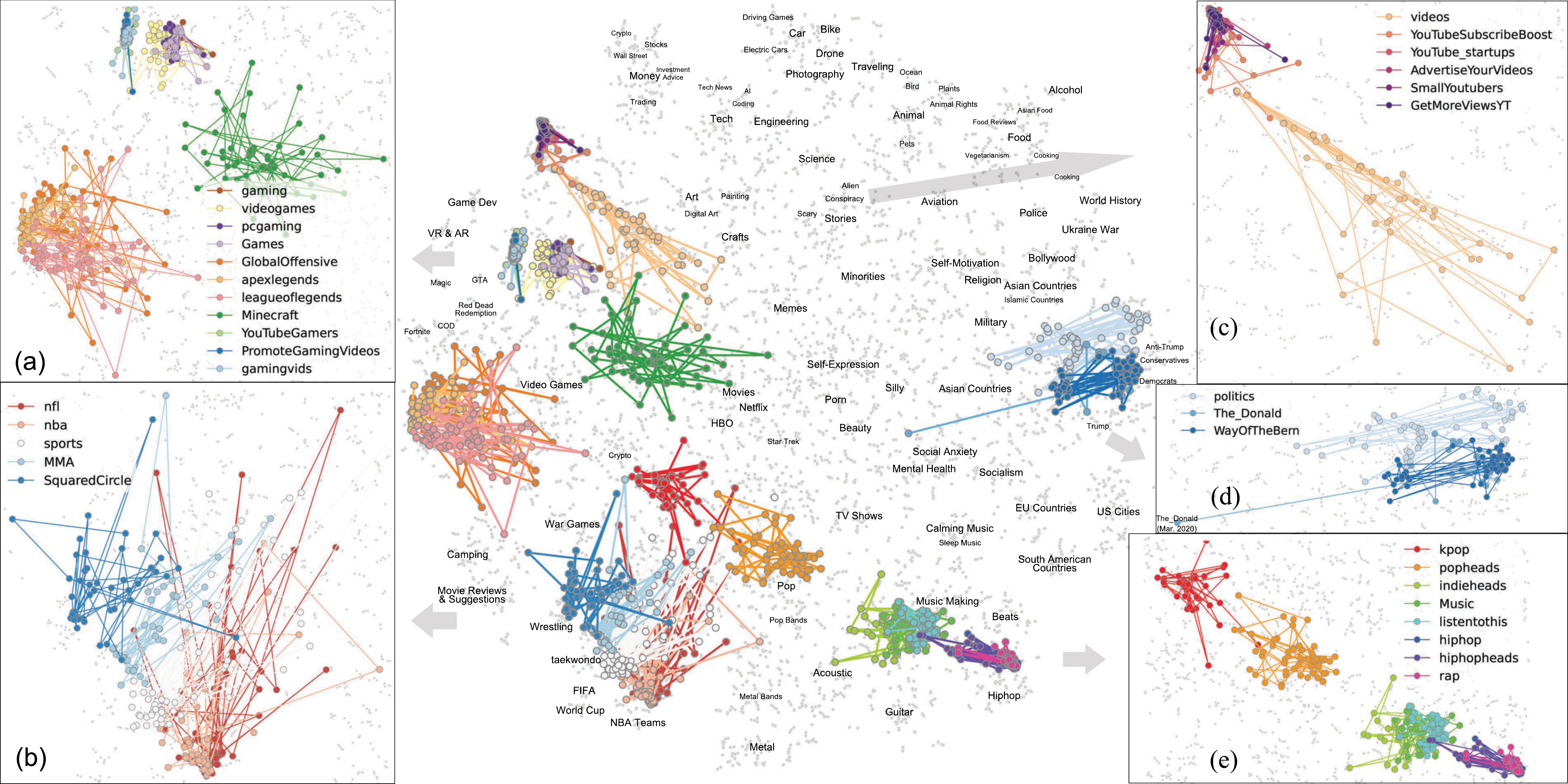}
    \caption{Visualization of Reddit online communities. Each gray node in the background represents an online community (``subreddit''). The trajectories of five groups of subreddits are displayed, including \textbf{a.} gaming, \textbf{b.} sports, \textbf{c.} video-sharing, \textbf{d.} politics, and \textbf{e.} music. Text in the background indicates the topics that characterize each subreddit cluster. Different video-sharing communities (\textbf{c.}) manifest diverse levels of specialization, where communities with a narrow focus of video promotion demonstrate less mobility than general-purpose communities. \model captures a major event in r/The\_Donald -- its shutdown.}
    \label{fig:reddit}
    \vspace{-0.5cm}
\end{figure*}
Online users often form communities around shared interests, beliefs, ethnicity, and geographical locations~\cite{jin2023predicting}. 
A deeper understanding of these community dynamics on platforms like \href{https://www.reddit.com/}{Reddit}, which is structured into thousands of interest-specific ``subreddits'', is crucial for analyzing how user groups interact, share content, and influence one another over time. 
This study presents an analysis of subreddit trajectories across various topics, including gaming, sports, videos, politics, and music, with a focus on content specialization and the phenomenon of echo chambers, as shown in Fig.~\ref{fig:reddit}. 
Each subreddit's trajectory is highlighted in a distinct color. To derive the graph embeddings, we train the model on the bipartite graph consisting of videos and subreddits, where an edge with timestamp $t$ exists between a video and a subreddit if the video is shared in the subreddit at $t$. 
In the resulting graph, two nodes are close in the embedding space if they share similar videos. 
Each subreddit's trajectory within the visualized graph embeddings indicates the level of content homogeneity or diversity. 

\begin{figure}[ht]
    \centering
    \includegraphics[width=0.95\textwidth]{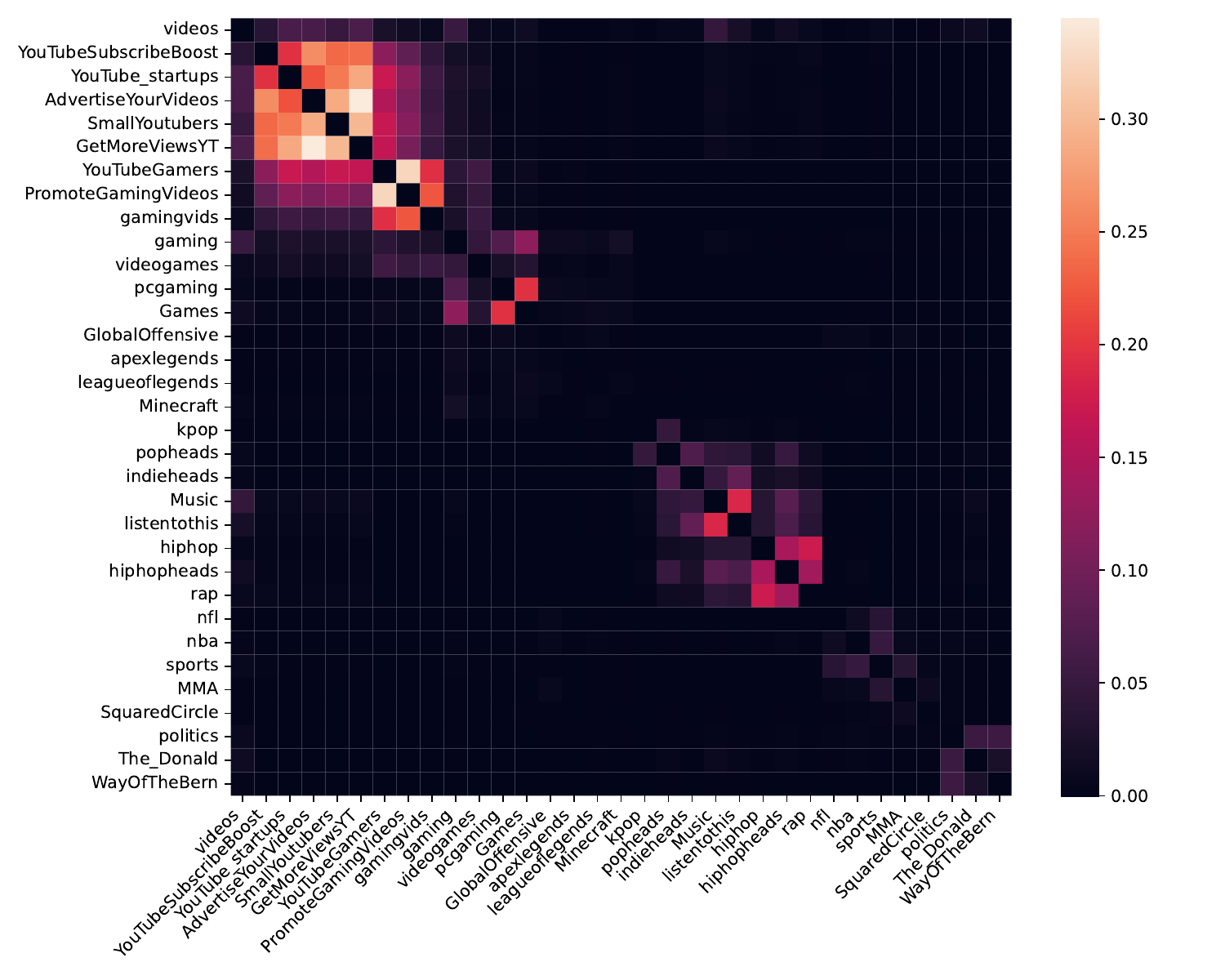}
    \caption{$\operatorname{Jaccard}_{100}$ among the subreddits in terms of videos shared between them.}
    \label{fig:RedditJaccard}
\end{figure}

\paragraph{Specialization in Content Sharing Across Video-Related Subreddits}
The trajectories of video-sharing subreddits (Fig.~\ref{fig:reddit}c) demonstrate diverse levels of specialization. Subreddits with a narrow focus on promoting YouTube videos and small channels, such as \texttt{r/GetMoreViewsYT}, r/YouTube\_startups, r/AdvertiseYourVideos, r/SmallYoutubers, and r/YouTubeSubscribeBoost, move within a confined region, illustrating a high degree of content homogeneity within these subreddits as users simultaneously spread the same videos within multiple subreddits for better visibility. In contrast, general subreddits like r/videos display a greater diversity of content, as shown from their more expansive trajectories. 
These findings are supported by the numeric values of $\mathrm{Jaccard}_{100}$ (Fig.~\ref{fig:RedditJaccard}), where the overlap between the nearest neighbors of each video-related subreddit in the embedding space in adjacent timestamps is high for video-promotion subreddits. Details for the metrics are in Supplementary Sec.~\ref{sec:analytics}. 

\paragraph{Diversity and Overlap in Sports-Related Subreddits} On the other hand, for sports-related communities (Fig.~\ref{fig:reddit}b), subreddits with specialized topics, such as r/nba (subreddit for the National Basketball Association), r/nfl (subreddit for the National Football League), r/MMA (subreddit for mixed martial arts), r/SquaredCircle (subreddit for professional wrestling), demonstrate similar levels of movements to more general subreddits like r/sports. 
Notably, r/nba has a large overlap with r/nfl, indicating that these two subreddits share similar audience, posts, and content sharing pattern. Both NBA and NFL feature team-based sports, high-profile athletes, strategies, and have regular seasons followed by playoff rounds that culminate in a championship event. 
In case of content sharing, many videos feature athletes or moments that have transcended their respective sports and gained widespread popularity, which is appreciated by fans of both basketball (NBA) and American football (NFL). 
Compared to subreddits focused on videos, sports-related subreddits display more variability among their neighboring subreddits in the embedding space, as evidenced by the $\operatorname{Jaccard}_{100}$ index values averaged on an annual basis (Table~\ref{tab:reddit}). 

\paragraph{Trajectories of Political Subreddits Reveal Echo Chamber and Major Events} 
The phenomenon of echo chambers within online social networks, wherein users experience reinforcement of their ideologies through repeated interactions with like-minded peers and a narrow spectrum of information, presents a significant challenge to discourse diversity~\cite{monti2021learning, del2016echo, cinelli2021echo}. 
This pattern is notably pervasive on platforms like Reddit, where close-knit communities form around specific ideologies or interests. 
A pertinent example is observed in the subreddit r/WayOfTheBern, an unofficial subreddit established by Bernie Sanders' supporters following his loss in the 2016 primary election~\cite{WayOfTheBernWashingtonTimes}. 
Initially intended as a space for political discourse divergent from the mainstream Democratic Party narrative, this community has been scrutinized for its alignment and user overlap with right-leaning communities, suggesting a complex web of ideological positioning that transcends conventional political boundaries~\cite{WayOfTheBernRedditpedia, WayOfTheBernWashingtonTimes}. 
The embedding trajectories in Fig.~\ref{fig:reddit}d reveals substantial connections between r/WayOfTheBern and r/The\_Donald, another banned subreddit known for sharing misinformation and controversial content~\cite{WayOfTheBernWashingtonTimes, mann2023unsorted}. 
These communities demonstrate converging paths that deviate from more generalized political forums like r/politics. 
Significantly, r/The\_Donald manifests a notable divergence in the its trajectory around March 2020, coinciding with key external events such as the COVID-19 outbreak and subsequent quarantine in US major cities. 
The trajectory of r/The\_Donald terminates at a juncture markedly distinct from its typical position in March 2020, coinciding with the outbreak of COVID-19 pandemic in the United States, the implementation of quarantine measures in major US cities, and Reddit's decision to relegate r/The\_Donald to ``Restricted mode'' and restricting most users from creating new posts~\cite{RedditClosedTheDonald}. Such a confluence of events indicates a notable divergence and deterioration characterized by the proliferation of toxic discourse within the community.

The existence and perpetuation of echo chambers underscore the complex challenges of online social networks in fostering balanced and open discourse. They not only facilitate the entrenchment of partisan beliefs by insulating users from contrary viewpoints but also serve as fertile grounds for the spread of misinformation. The observed patterns and trajectories within these communities highlight the urgent need for strategies aimed at early detection and mitigation of echo chambers, ensuring a more diverse and accurate exchange of information within these digital ecosystems.





\subsection{Linguistic Reflections and Shifts in Societal Perceptions Through Lexicon Graphs}
\label{sec:HistWords}


Semantic shifts in word meanings usually reveal socio-cultural changes over time, 
whereas the rates of semantic change vary significantly across words~\cite{hamilton2016diachronic}. By leveraging word embeddings, \model effectively tracks the dynamics of lexical connotations over time. 
Our study uses the skip-gram with negative sampling (SGNS) embeddings~\cite{hamilton2016diachronic} trained on the Google N-Gram~\cite{lin2012syntactic} dataset. 
Conventional SGNS approach typically considers a fixed window of context words around the target word, and thus may not fully capture the contextual meaning of a word, especially in intricate linguistic contexts characterized by long-range dependencies or when dealing with semantically similar words with limited co-occurrence within the local context.
To overcome this issue, we construct a new temporal graph. For each timestamp $t$, we compute the pairwise cosine similarity between each pair of word embeddings $\mathbf{v}_i^t$ and $\mathbf{v}_j^t$, and then connect each word to its $k$ nearest neighbors with the highest cosine similarity. A new set of temporal embeddings is then trained on this graph. This method facilitates the extraction of high-order semantic associations between words that may not typically co-occur within the same local context, thus overcoming the limitation of the original SGNS embeddings. 
Empirically, we experimented with $k \in [5, 10, 20, 50, 100]$, and found that $k=20$ yields the most meaningful semantic associations between words. 

\begin{figure*}
    \centering    \includegraphics[width=0.98\textwidth]{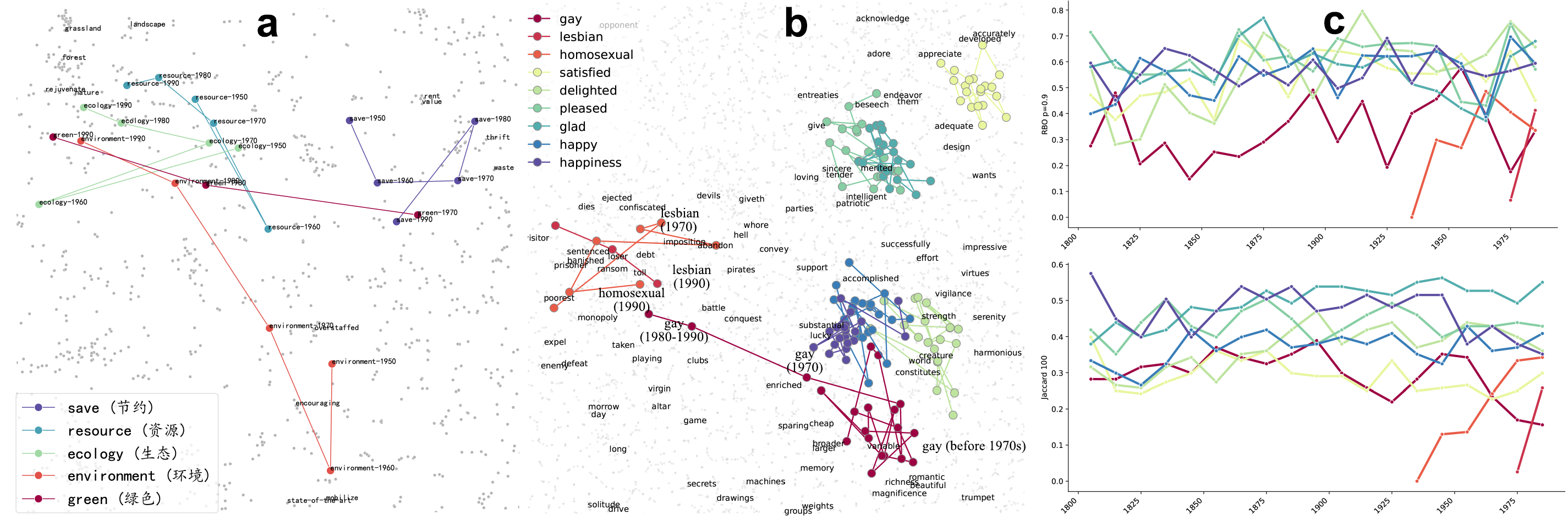}
    \caption{\textbf{a.} 
    Chronological evolution of Chinese lexicon pertaining to environmental protection. 
    These words exhibit diverse meanings from the 1950s onward, culminating in a cohesive cluster by the 1990s.
    This trend underscores the growing prominence and consolidation of environmental protection concepts within the analyzed corpus. 
    English translations are provided for reference. \textbf{b.} Semantic Shift in LGBTQ+ Terminology. 
    The word ``gay'' was initially synonymous with ``joy'' and ``happiness,'' but its usage progressively aligns with homosexuality. This shift underscores the changing societal discourse and recognition of LGBTQ+ identities. 
    \textbf{c}. Comparative analysis of semantic stability using RBO and $\textrm{Jaccard}_{100}$ reveal that words related to homosexuality exhibit substantial shifts in meaning since their inception, reflecting societal changes in perception and language. In contrast, terms solely associated with happiness show remarkable semantic stability, highlighting the enduring nature of certain lexicons despite evolving societal contexts. }
    \label{fig:HistWordsEN}
\end{figure*}

\paragraph{Tracing the Evolution of Socio-Economic Language in Environmental Discourse from 1950s to 1990s}
To gain insights into the evolving socio-economic discourse concerning environmental concerns, we delve into the semantic trajectories of words associated with environmental protection using the HistWords-CN dataset (Fig.~\ref{fig:HistWordsEN}a). 
Starting from the 1950s, we traced diverse interpretations of words such as ``environment,'' which initially carried connotations related to the working environment, as indicated by their proximity to words like ``team,'' ``mobilization,'' and ``state-of-the-art.'' 
However, as we move into the 1980s and 1990s, we observe a convergence of these terms toward the region occupied by ecological-environment-related words, such as ``forest,'' ``grassland,'' ``carbon dioxide,'' and ``nature.'' 
This reflects the ever-growing discourse on ecology and the escalating importance attached to environmental protection. Notably, despite this convergence, the term ``save'' deviated from this trajectory due to its diverse meanings related to cost-saving, rent, thrift, and value. Our model thus provides an intricate understanding of the evolution of environmental discourse over time.


\subsection{Evolving Language and LGBTQ+ Acceptance: A Lexical Analysis of Societal Shifts}

The power of language lies in its ability to both reflect and shape societal attitudes. In this context, we explore the linguistic landscape surrounding homosexuality, recognizing its historical significance as a mirror for societal changes. 

\paragraph{The term ``Gay'' Experienced Significant Lexical Evolution in the 1970s.}
During the 1970s, the LGBTQ+ rights movement in the United States experienced a transformative period characterized by increased visibility and activism~\cite{moore2004beyond}. However, prevailing societal attitudes during this era remained heavily influenced by traditional values and social norms, often stigmatizing homosexuality~\cite{cohen2007gay}. 
As depicted in Fig.~\ref{fig:HistWordsEN}, we observe a remarkable lexical shift associated with the term ``gay'' from the 1970s to the 1990s. The word gradually transitions away from its original connotations of happiness and fortune towards homosexuality, aligning with its etymological evolution~\cite{jatowt2014framework}. 

Additionally, Table~\ref{tab:gaywords} provides a comprehensive view of the top five words associated with each term in the embedding space over time. The words ``happy'' and ``delighted'' retain their consistent meanings across the years, serving as constants in the lexical landscape. However, the term ``gay,'' once widely employed to convey happiness before the 1960s, underwent a profound transformation when they were used to refer to homosexuality in the 1970s and acquired proximity with negative words such as ``forlorn'' and ``ugly.'' This lexical shift reflects the societal struggle to grapple with evolving perceptions of homosexuality. 
Furthermore, LGBT-related words, including ``gay,'' ``homosexual,'' and ``lesbian,'' exhibit strong associations with ``clubs'' and ``dance'' during the 1970s and 1980s. This phenomenon corresponds to the development of a distinctive LGBTQ+ culture and language during this era. Bars and dance clubs emerged as vital meeting places for the LGBTQ+ community, providing safe spaces for socialization, self-expression, and the formation of supportive networks~\cite{lusby2011ghent}.  It is crucial to acknowledge that the portrayal of LGBTQ+ characters and issues in popular culture largely perpetuated negative stereotypes and discriminatory portrayals during the examined period. This further entrenched negative attitudes within the general public, making societal acceptance and understanding a complex and arduous journey~\cite{mcinroy2017perspectives, nadal2016microaggressions}.

By meticulously tracing these linguistic transformations and contextualizing them within historical and societal frameworks, our study contributes to a deeper understanding of the intricate relationship between language, societal attitudes, and the ongoing struggle for LGBTQ+ acceptance.


\begin{figure*}[ht]
    \centering
    \includegraphics[width=0.95\textwidth]{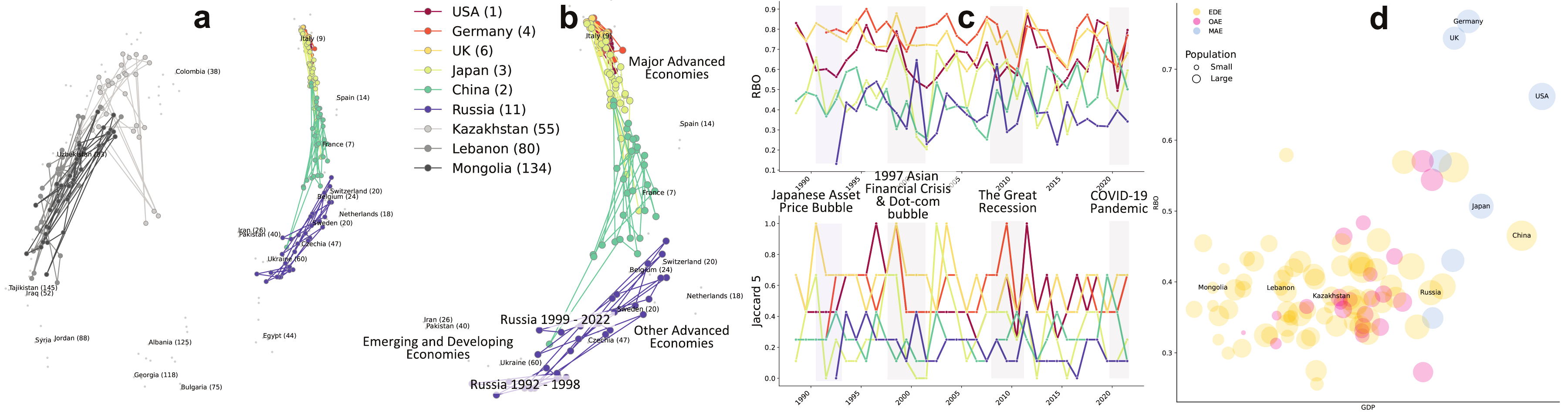} 
    \caption{\textbf{a.} Overview of the embedding trajectories on UN Comtrade~\cite{comtrade2010united}. Each country is labeled with its nominal GDP rankings in 2017~\cite{GDP2017} (\eg,``USA (1)''). Large- and middle-scale economies (\eg, USA, UK, Russia, Netherlands) with higher GDP rankings and intensive trade relations form a distinct cluster, while lower-ranked economies (\eg, Tajikistan, Uzbekistan, Jordan) exhibit individual clusters; 
    \textbf{b.} Detailed view of \textbf{a}. The three country groups according to IMF~\cite{IMFCountryGroups}, Major Advanced Economies (MAE), Other Advanced Economies (OAE), and Emerging and Developing Economies (EDE), form distinct visual partitions. The trajectories of advanced economies with economic stability, such as the USA, UK, and Germany, remain in a constrained region, while countries that have experienced rapid growth or drastic economic instability, such as Japan, China, and Russia, manifest more diverse trajectories; 
    \textbf{c.} Fluctuations in $\operatorname{Jaccard}_n$ and RBO align with major economic events in history. 
    \textbf{d.} Average RBO of each country over the period 1988-2022. The x-axis describes the total GDP on a logarithmic scale. Node colors indicate country types, and node sizes represent the population. 
    }
    \label{fig:UNComtrade}
\end{figure*}
\subsection{Unveiling Global Trade Dynamics: Insights from UN Comtrade Export Data. }
\label{sec:UNComtrade}
In the field of economics, understanding, modeling, and predicting international trade plays a crucial role in helping economists and policymakers navigate the challenges and opportunities arising from globalization, such as financial crises. In this study, we analyze international trade dynamics using export data from the United Nations Commodity Trade Statistics Database~\cite{comtrade2010united}.
To capture the economic status and trading partnerships of countries, we perform linear regression on the logarithmic values of a country's gross exports and the bilateral trade volumes, and employ the joint training objective with $\lambda_1=\lambda_2=0.1$ in Equation~\ref{eq:totalLoss}. 

The resulting visualization in Fig.~\ref{fig:UNComtrade}a offers a comprehensive representation of the international trade landscape. 
Advanced Economies, as classified by the International Monetary Fund (IMF)\footnote{\url{https://www.imf.org/en/Publications/WEO/weo-database/2023/April/groups-and-aggregates}}, form distinct clusters located primarily in the upper right region, while countries with lower trade volumes and those positioned on the periphery of international trade form separate clusters in the left and lower regions. 
In addition, Fig.~\ref{fig:UNComtrade}b provides a clearer illustration of the distinct visual partitions among the three country groups defined by IMF~\cite{IMFCountryGroups}: Major Advanced Economies (MAE)\footnote{IMF defines ``Major Advanced Economies'' as the G7 countries, including Canada, France, Germany, Italy, Japan, the UK and the USA}, Other Advanced Economies (OAE), and Emerging and Developing Economies (EDE). This spatial arrangement reflects the different degrees of trade engagement of each country within the global trade network. 

\paragraph{Dynamic Graph Embedding Trajectories of Individual Countries Reveal Development and Stability Patterns of Key Economies} In Fig.~\ref{fig:UNComtrade}b, the trajectories of individual countries reveal distinct patterns of economic development and stability. The United States, the United Kingdom, and Germany\footnote{Germany has been listed in UN Comtrade as a single sovereign state since 1991, following German reunification in October 1990.} demonstrate relatively stable and consistent trading status throughout the examined period (1988-2022). 
On the other hand, China's trajectory moves between MAE and OAE, reflecting its prolonged period of economic development characterized by comprehensive domestic reforms, the lifting of price controls, and the liberalization of trade policies~\cite{longworth2001beef, lin2004china}. 
Russia exhibits significant movements between OAE and EDE. Its trajectory predominantly shifted towards the EDE region during the period 1992-1998, coinciding with a substantial 40\% contraction in GDP~\cite{monken2021graph}. Starting from the early 2000s, Russia moves towards the region occupied by OAEs, including the four middle-sized developed countries Switzerland, Belgium, Sweden, and the Netherlands, which indicates a period of economic recovery characterized by greater trade volumes. 
Despite its status as an MAE, Japan has experienced economic development with significant fluctuations. The country encountered unique obstacles such as the Japanese asset price bubble (1990-1992) whose impact has lasted for more than a decade~\cite{JapanBubbleEconomy, okina2001asset}. 
We further use the Jaccard index~\cite{jaccard1912distribution} and Rank-biased Overlap (RBO)~\cite{webber2010similarity, oh2022rank} to measure the macro-level changes over time. Detailed calculations of these metrics are in Supplementary Sec.~\ref{sec:analytics}. 
As reflected in Fig.~\ref{fig:UNComtrade}c, the RBO and $\mathrm{Jaccard}_{5}$ for Japan plummeted during this period compared to other countries, indicating a period of instability in its economic status. 

\paragraph{Trade Resilience and Volatility during Global Economic Crises}
From Fig.~\ref{fig:UNComtrade}c, we observe three periods of significant fluctuations in RBO and $\mathrm{Jaccard}_{5}$ for most countries, indicating significant changes in their trading status. 
The first period is 1997 - 2003, which corresponds to the 1997 Asian Financial Crisis and the dot-com bubble when investor confidence declined worldwide. 
For most countries, the recovery from the financial crisis in 1998–1999 was rapid~\cite{radelet1998east}. For example, China demonstrates quick movements towards and away from the EDE region (Fig.~\ref{fig:UNComtrade}b) around 1998. 
These two events had global ripple effects. 
The dot-com bubble, during which many large-scale Internet and communication companies failed and shut down, has a more far-reaching effect.
As the epicenter of the bubble, the US experienced the most drastic fluctuation in its trading status, as shown by its decline in RBO and $\operatorname{Jaccard}_5$~\cite{jaccard1912distribution} in Fig.~\ref{fig:UNComtrade}c (Refer to appendix for ).
Similarly, the Great Recession in the 2008s and the COVID-19 also caused fluctuations in RBO and $\text{Jaccard}_5$. 

\subsection{Dynamic Graph Analysis of Gene Expression Trajectories Reveals Key Patterns in Aging}
Dynamic graphs are vital for identifying anomalous genes and genetic variations that significantly impact disease development~\cite{akoglu2015graph} and human aging process~\cite{huang2022dgraph}. 
\model enables researchers to effectively pinpoint genes with unusual patterns or interactions, facilitating a deeper understanding of aging-related diseases, the genetic mechanisms underlying the aging process, and potential treatments. 
\begin{figure*}[ht]
    \centering
    \includegraphics[width=0.95\textwidth]{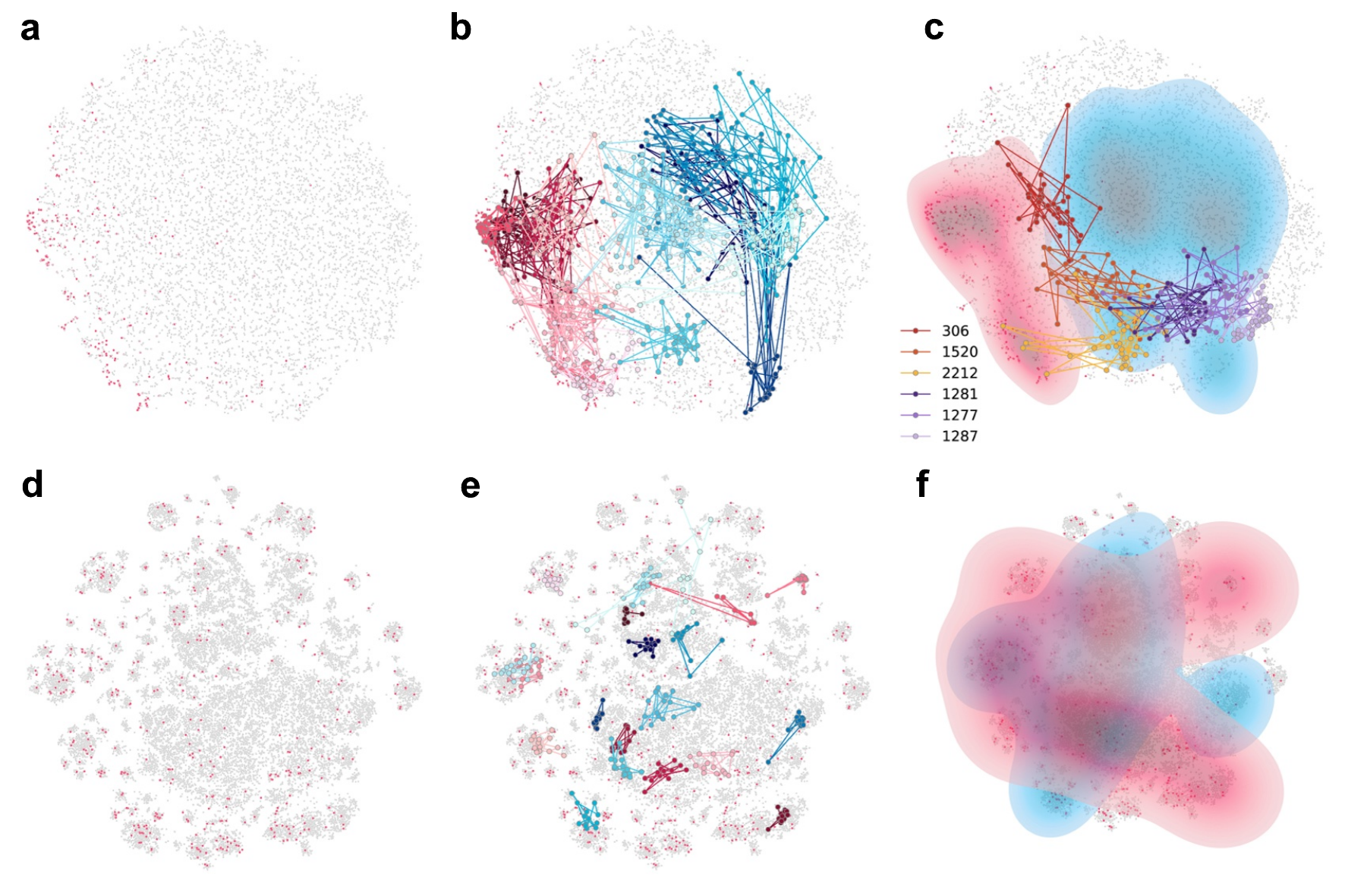} 
    \caption{\textbf{a/d.} t-SNE visualization of the Aging dataset~\cite{li2021improved} and the DGraph dataset~\cite{huang2022dgraph}. Red dots represent aging-related genes in the genetic network (\textbf{a}) and fraudsters in the financial network (\textbf{d}). Gray dots represent normal nodes (non-aging-related genes and normal users), respectively. \textbf{b/e.} embedding trajectories of 10 anomalous nodes (in warm colors) and 10 normal nodes (in cold colors), respectively \textbf{c/f.} The kernel density estimate (KDE) plot for the trajectories. Darker colors indicate higher node densities.}
    \label{fig:Aging}
    \vspace{-0.5cm}
\end{figure*}
\paragraph{Characterizing Structural and Temporal Differences in Gene Expression During Aging} We examine structural differences, neighbor distributions, and temporal dynamics between aging-related and non-aging-related genes using human gene expression data at 37 differnt ages, ranging from 20 to 99. 
The t-SNE projection in Fig.~\ref{fig:Aging}a shows that genes \emph{directly related to aging} (red dots) have distinct distributions from normal genes (gray dots). 
We further analyze the trajectories of 10 aging-related and 10 non-aging-related genes, and plot their embedding trajectories in Fig.~\ref{fig:Aging}b. From a dynamic graph perspective, the aging-related genes are characterized by distinct embedding trajectories, which are mainly located on the left side of the plot. Such distinctions are reinforced by the kernel density estimation (KDE) plot in Fig.~\ref{fig:Aging}c. 

\paragraph{Application of \model for Predicting Aging-Related Gene Behavior} 
We randomly select 6 genes commonly altered during the human aging process as identified in previous research~\cite{tacutu2012human}. These genes experience frequent changes due to their roles in cellular processes, although there is insufficient evidence linking them directly to aging. These genes are categorized as overexpressed (Gene 306, 1520, and 2212) and underexpressed genes (Gene 1281, 1277, and 1287)~\footnote{\url{https://genomics.senescence.info/genes/microarray.php}}. 
As shown in Fig.~\ref{fig:Aging}c, the orange trajectories representing overexpressed genes typically transition between regions associated with aging and non-aging, suggesting that these genes can potentially induce or accelerate the aging process, despite the absence concrete evidence. 
Meanwhile, the purple trajectories representing underexpressed genes mostly remaining within non-aging regions, suggesting that these genes are less likely to be involved in the aging process. 


\subsection{Challenges in Distinguishing Fraudulent and Legitimate Behaviors in Financial Networks}

Dynamic graphs can be used in financial networks to detect and flag users engaged in fraudulent behaviors~\cite{huang2022dgraph}. Accurate identification of fraudulent users can facilitate timely intervention and prevent financial loss. 
As shown in Fig.~\ref{fig:Aging}e, the distinction between fraudsters and normal users appears less pronounced, as both groups exhibit trajectories widely dispersed across the plot. 
These observations highlight the challenge of distinguishing between fraudulent and normal users. 
In real-world scenarios, fraudulent users possess a remarkable ability to camouflage their activities, often mirroring the behaviors of genuine users. 
This challenge is further exemplified in Fig.~\ref{fig:Aging}f, where the KDE plot depicts the convergence of their trajectories, 
underscoreing the complexity in accurately identifying and differentiating fraudulent activities from legitimate ones. 

\subsection{Modeling Social Dynamics in Ant Colonies on Animal Activity Graphs}

Animals exhibit intricate and efficient social organizations. For example, ant colonies demonstrate as well-defined organizational hierarchy and role differentiation among worker ants~\cite{mersch2013tracking}. 
Roles within these societies include nurses, responsible for the care of the brood and the queen; cleaners, who ensure colony cleanliness and waste disposal; and foragers, tasked with acquiring food resources from outside the colony.
Dynamic graph modeling is utilized to describe these behaviors and the evolution of social roles within animal groups. Our model provides a clear interpretation of the trajectories of role-based behaviors, as inferred from the embedding model. 

\begin{figure}[ht]
    \centering
    \includegraphics[width=0.5\textwidth]{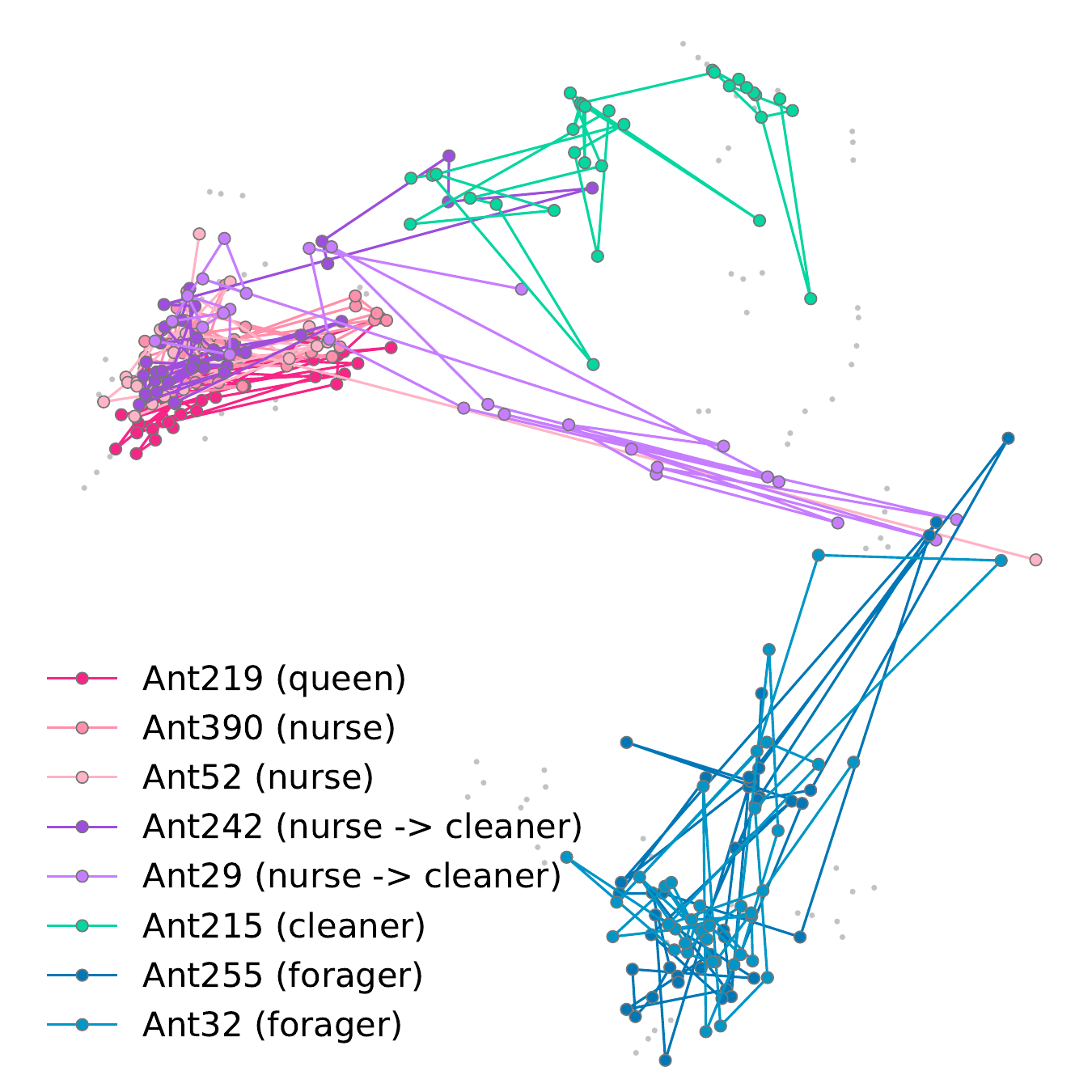}
    \caption{Trajectories of ants from the Ant dataset~\cite{mersch2013tracking}. }
    \label{fig:ant}
    \vspace{-0.2cm}
\end{figure}

\paragraph{Trajectories of Different Ant Roles Reveal Distinct Spatial Organizations} Fig.~\ref{fig:ant} illustrates these findings, showing that the movement patterns of nurses are generally restricted to areas near the queen, reflecting their frequent interactions. Conversely, foragers are typically found in remote areas, aligning with their external foraging activities and minimal contact with the queen. The spatial distribution and movements of these roles over time reveal distinct patterns: nurses and foragers maintain localized activity areas, whereas cleaners exhibit movement patterns intersecting with those of nurses due to their intermediary tasks. 

\paragraph{Capturing Role Transitions in Ant Behaviors} \model captures the transition of individuals between roles, a phenomenon supported by existing literature~\cite{mersch2013tracking}. For example, the trajectories of certain ants (e.g., Ant29 and Ant242) shift from nursing towards cleaning roles over time, indicating a natural progression as they age. This dynamic is effectively represented in our models
providing insight into the adaptive behaviors within ant colonies. 


\section{Conclusion and Future Works}
\label{sec:conclusion}
In this work, we formally define the problem of \problem, and introduce \model, a novel framework to effectively address the problem. 
Empirical evaluation on 9 real-world datasets demonstrates the broad application of \model and provides significant insights. 

Looking forward, there are multiple promising directions for further research. An immediate area of interest is the development of more refined methodologies for assessing the quality and efficacy of the visualizations generated. This could involve the creation of metrics and evaluation protocols that better capture the utility and interpretability of visual outputs in practical scenarios. Additionally, it is imperative to investigate the potential of \model to be adapted or enhanced to support a wider array of visualization paradigms and representations. Such explorations could extend its relevance to other data types and structures beyond graphs, thereby accommodating the dynamic and diverse needs of modern data visualization. 

\section*{Acknowledgment}
\label{sec:acknowledgement}
This research/material is based upon work supported in part by NSF grants CNS-2154118, IIS-2027689, ITE-2137724, ITE-2230692, CNS2239879, Defense Advanced Research Projects Agency (DARPA) under Agreement No. HR00112290102 (subcontract No. PO70745), and funding from Microsoft, Google, and Adobe Inc. Any opinions, findings, and conclusions or recommendations expressed in this material are those of the author(s) and do not necessarily reflect the position or policy of DARPA, DoD, SRI International, NSF and no official endorsement should be inferred. We thank the reviewers for their comments.

\bibliographystyle{unsrt}
\bibliography{cite}

\begin{thebibliography}{10}

\bibitem{jin2023predicting}
Yiqiao Jin, Yeon-Chang Lee, Kartik Sharma, Meng Ye, Karan Sikka, Ajay
  Divakaran, and Srijan Kumar.
\newblock Predicting information pathways across online communities.
\newblock In {\em KDD}, 2023.

\bibitem{hamilton2016diachronic}
William~L Hamilton, Jure Leskovec, and Dan Jurafsky.
\newblock Diachronic word embeddings reveal statistical laws of semantic
  change.
\newblock In {\em ACL}, pages 1489--1501, 2016.

\bibitem{monken2021graph}
Anderson Monken, Flora Haberkorn, Munisamy Gopinath, Laura Freeman, and Feras~A
  Batarseh.
\newblock Graph neural networks for modeling causality in international trade.
\newblock In {\em FLAIRS}, volume~34, 2021.

\bibitem{huang2022dgraph}
Xuanwen Huang, Yang Yang, Yang Wang, Chunping Wang, Zhisheng Zhang, Jiarong Xu,
  Lei Chen, and Michalis Vazirgiannis.
\newblock Dgraph: A large-scale financial dataset for graph anomaly detection.
\newblock {\em NIPS}, 35:22765--22777, 2022.

\bibitem{you2022roland}
Jiaxuan You, Tianyu Du, and Jure Leskovec.
\newblock Roland: graph learning framework for dynamic graphs.
\newblock In {\em KDD}, pages 2358--2366, 2022.

\bibitem{pareja2020evolvegcn}
Aldo Pareja, Giacomo Domeniconi, Jie Chen, Tengfei Ma, Toyotaro Suzumura,
  Hiroki Kanezashi, Tim Kaler, Tao Schardl, and Charles Leiserson.
\newblock Evolvegcn: Evolving graph convolutional networks for dynamic graphs.
\newblock In {\em AAAI}, volume~34, pages 5363--5370, 2020.

\bibitem{seo2018structured}
Youngjoo Seo, Micha{\"e}l Defferrard, Pierre Vandergheynst, and Xavier Bresson.
\newblock Structured sequence modeling with graph convolutional recurrent
  networks.
\newblock In {\em ICONIP}, pages 362--373. Springer, 2018.

\bibitem{li2020argo}
Siwei Li, Zhiyan Zhou, Anish Upadhayay, Omar Shaikh, Scott Freitas, Haekyu
  Park, Zijie~J Wang, Susanta Routray, Matthew Hull, and Duen~Horng Chau.
\newblock Argo lite: Open-source interactive graph exploration and
  visualization in browsers.
\newblock In {\em CIKM}, pages 3071--3076, 2020.

\bibitem{mersch2013tracking}
Danielle~P Mersch, Alessandro Crespi, and Laurent Keller.
\newblock Tracking individuals shows spatial fidelity is a key regulator of ant
  social organization.
\newblock {\em Science}, 340(6136):1090--1093, 2013.

\bibitem{hotelling1933analysis}
Harold Hotelling.
\newblock Analysis of a complex of statistical variables into principal
  components.
\newblock {\em Journal of educational psychology}, 24(6):417, 1933.

\bibitem{van2008visualizing}
Laurens Van~der Maaten and Geoffrey Hinton.
\newblock Visualizing data using t-sne.
\newblock {\em JMLR}, 9(11), 2008.

\bibitem{pezzotti2016hierarchical}
Nicola Pezzotti, Thomas H{\"o}llt, B~Lelieveldt, Elmar Eisemann, and Anna
  Vilanova.
\newblock Hierarchical stochastic neighbor embedding.
\newblock In {\em Computer Graphics Forum}, volume~35, pages 21--30. Wiley
  Online Library, 2016.

\bibitem{an2020viva}
Sungtae An, Shenda Hong, and Jimeng Sun.
\newblock Viva: semi-supervised visualization via variational autoencoders.
\newblock In {\em ICDM}, pages 22--31. IEEE, 2020.

\bibitem{chen2022gc}
Jinyin Chen, Xueke Wang, and Xuanheng Xu.
\newblock Gc-lstm: Graph convolution embedded lstm for dynamic network link
  prediction.
\newblock {\em Applied Intelligence}, pages 1--16, 2022.

\bibitem{sankar2020dysat}
Aravind Sankar, Yanhong Wu, Liang Gou, Wei Zhang, and Hao Yang.
\newblock Dysat: Deep neural representation learning on dynamic graphs via
  self-attention networks.
\newblock In {\em WSDM}, pages 519--527, 2020.

\bibitem{jaccard1912distribution}
Paul Jaccard.
\newblock The distribution of the flora in the alpine zone. 1.
\newblock {\em New phytologist}, 11(2):37--50, 1912.

\bibitem{webber2010similarity}
William Webber, Alistair Moffat, and Justin Zobel.
\newblock A similarity measure for indefinite rankings.
\newblock {\em TOIS}, 28(4):1--38, 2010.

\bibitem{oh2022rank}
Sejoon Oh, Berk Ustun, Julian McAuley, and Srijan Kumar.
\newblock Rank list sensitivity of recommender systems to interaction
  perturbations.
\newblock In {\em CIKM}, pages 1584--1594, 2022.

\bibitem{monti2021learning}
Corrado Monti, Giuseppe Manco, Cigdem Aslay, and Francesco Bonchi.
\newblock Learning ideological embeddings from information cascades.
\newblock In {\em CIKM}, pages 1325--1334, 2021.

\bibitem{del2016echo}
Michela Del~Vicario, Gianna Vivaldo, Alessandro Bessi, Fabiana Zollo, Antonio
  Scala, Guido Caldarelli, and Walter Quattrociocchi.
\newblock Echo chambers: Emotional contagion and group polarization on
  facebook.
\newblock {\em Scientific reports}, 6(1):37825, 2016.

\bibitem{cinelli2021echo}
Matteo Cinelli, Gianmarco De~Francisci~Morales, Alessandro Galeazzi, Walter
  Quattrociocchi, and Michele Starnini.
\newblock The echo chamber effect on social media.
\newblock {\em PNAS}, 118(9):e2023301118, 2021.

\bibitem{WayOfTheBernWashingtonTimes}
James Varney.
\newblock Prominent pro-sanders subreddit wayofthebern aims to divide
  democrats, says social media analyst.
\newblock {\em The Washington Times}, 2 2019.

\bibitem{WayOfTheBernRedditpedia}
Redditpedia Wiki.
\newblock Subreddit statistics of user overlap, 2023.

\bibitem{mann2023unsorted}
Marcus Mann, Diana Zulli, Jeremy Foote, Emily Ku, and Emily Primm.
\newblock Unsorted significance: Examining potential pathways to extreme
  political beliefs and communities on reddit.
\newblock {\em Socius}, 9:23780231231174823, 2023.

\bibitem{RedditClosedTheDonald}
Elizabeth Timberg, Craig;~Dwoskin.
\newblock Reddit closes long-running forum supporting president trump after
  years of policy violations.
\newblock {\em The Washington Post}, 2020.

\bibitem{lin2012syntactic}
Yuri Lin, Jean-Baptiste Michel, Erez~Aiden Lieberman, Jon Orwant, Will
  Brockman, and Slav Petrov.
\newblock Syntactic annotations for the google books ngram corpus.
\newblock In {\em ACL}, pages 169--174, 2012.

\bibitem{moore2004beyond}
Patrick Moore.
\newblock {\em Beyond shame: Reclaiming the abandoned history of radical gay
  sexuality}.
\newblock Beacon Press, 2004.

\bibitem{cohen2007gay}
Stephan Cohen.
\newblock {\em The Gay Liberation Youth Movement in New York:'an army of lovers
  cannot fail'}.
\newblock Routledge, 2007.

\bibitem{jatowt2014framework}
Adam Jatowt and Kevin Duh.
\newblock A framework for analyzing semantic change of words across time.
\newblock In {\em JCDL}, pages 229--238. IEEE, 2014.

\bibitem{lusby2011ghent}
Michael~Anthony Lusby.
\newblock Ghent gayland: A case study of the gay and lesbian community and
  media of norfolk, virginia.
\newblock Master's thesis, College of William \& Mary, 2011.

\bibitem{mcinroy2017perspectives}
Lauren~B McInroy and Shelley~L Craig.
\newblock Perspectives of lgbtq emerging adults on the depiction and impact of
  lgbtq media representation.
\newblock {\em Journal of youth studies}, 20(1):32--46, 2017.

\bibitem{nadal2016microaggressions}
Kevin~L Nadal, Chassitty~N Whitman, Lindsey~S Davis, Tanya Erazo, and Kristin~C
  Davidoff.
\newblock Microaggressions toward lesbian, gay, bisexual, transgender, queer,
  and genderqueer people: A review of the literature.
\newblock {\em The journal of sex research}, 53(4-5):488--508, 2016.

\bibitem{comtrade2010united}
UN~Comtrade.
\newblock The united nations commodity trade statistics database.
\newblock {\em https://comtrade.un.org/}, 2010.

\bibitem{GDP2017}
Worldometer.
\newblock Gdp by country (2017), 2023.

\bibitem{IMFCountryGroups}
IMF.
\newblock Country composition of weo groups, 2023.

\bibitem{longworth2001beef}
John~William Longworth, Colin~G Brown, and Scott~A Waldron.
\newblock Beef in china: agribusiness opportunities and challenges.
\newblock {\em The China Journal}, 2001.

\bibitem{lin2004china}
Justin~Yifu Lin, Fang Cai, and Zhou Li.
\newblock {\em The China miracle: Development strategy and economic reform
  (Revised Edition)}.
\newblock The Chinese University of Hong Kong Press, 2004.

\bibitem{JapanBubbleEconomy}
Thayer Watkins.
\newblock Japan's bubble economy, 1999.

\bibitem{okina2001asset}
Kunio Okina, Masaaki Shirakawa, and Shigenori Shiratsuka.
\newblock The asset price bubble and monetary policy: Japan’s experience in
  the late 1980s and the lessons.
\newblock {\em Monetary and Economic Studies (special edition)},
  19(2):395--450, 2001.

\bibitem{radelet1998east}
Steven Radelet, Jeffrey~D Sachs, Richard~N Cooper, and Barry~P Bosworth.
\newblock The east asian financial crisis: diagnosis, remedies, prospects.
\newblock {\em Brookings papers on Economic activity}, 1998(1):1--90, 1998.

\bibitem{akoglu2015graph}
Leman Akoglu, Hanghang Tong, and Danai Koutra.
\newblock Graph based anomaly detection and description: a survey.
\newblock {\em TKDE}, 29:626--688, 2015.

\bibitem{li2021improved}
Qi~Li, Khalique Newaz, and Tijana Milenkovi{\'c}.
\newblock Improved supervised prediction of aging-related genes via weighted
  dynamic network analysis.
\newblock {\em BMC bioinformatics}, 22(1):1--26, 2021.

\bibitem{tacutu2012human}
Robi Tacutu, Thomas Craig, Arie Budovsky, Daniel Wuttke, Gilad Lehmann, Dmitri
  Taranukha, Joana Costa, Vadim~E Fraifeld, and Joao~Pedro De~Magalhaes.
\newblock Human ageing genomic resources: integrated databases and tools for
  the biology and genetics of ageing.
\newblock {\em Nucleic acids research}, 41(D1):D1027--D1033, 2012.

\bibitem{klimt2004introducing}
B~KLIMT.
\newblock Introducing the enron corpus.
\newblock In {\em CEAS}, 2004.

\bibitem{rozemberczki2021chickenpox}
Benedek Rozemberczki, Paul Scherer, Oliver Kiss, Rik Sarkar, and Tamas Ferenci.
\newblock Chickenpox cases in hungary: a benchmark dataset for spatiotemporal
  signal processing with graph neural networks.
\newblock {\em arXiv preprint arXiv:2102.08100}, 2021.

\bibitem{newaz2020inference}
Khalique Newaz and Tijana Milenkovi{\'c}.
\newblock Inference of a dynamic aging-related biological subnetwork via
  network propagation.
\newblock {\em TCBB}, 19(2):974--988, 2020.

\bibitem{szymanski2017temporal}
Terrence Szymanski.
\newblock Temporal word analogies: Identifying lexical replacement with
  diachronic word embeddings.
\newblock In {\em ACL}, pages 448--453, 2017.

\bibitem{dadu2023application}
Anant Dadu, Vipul~K Satone, Rachneet Kaur, Mathew~J Koretsky, Hirotaka Iwaki,
  Yue~A Qi, Daniel~M Ramos, Brian Avants, Jacob Hesterman, Roger Gunn, et~al.
\newblock Application of aligned-umap to longitudinal biomedical studies.
\newblock {\em Patterns}, 4(6), 2023.

\bibitem{mcinnes2018umap}
Leland McInnes, John Healy, Nathaniel Saul, and Lukas Gro{\ss}berger.
\newblock Umap: Uniform manifold approximation and projection.
\newblock {\em JOSS}, 3(29), 2018.

\bibitem{roweis2000nonlinear}
Sam~T Roweis and Lawrence~K Saul.
\newblock Nonlinear dimensionality reduction by locally linear embedding.
\newblock {\em science}, 290(5500):2323--2326, 2000.

\bibitem{tenenbaum2000global}
Joshua~B Tenenbaum, Vin~de Silva, and John~C Langford.
\newblock A global geometric framework for nonlinear dimensionality reduction.
\newblock {\em science}, 290(5500):2319--2323, 2000.

\bibitem{spearman1904proof}
C~Spearman.
\newblock The proof and measurement of association between two things.
\newblock {\em The American Journal of Psychology}, 15(1):72--101, 1904.

\bibitem{kendall1938new}
Maurice~G Kendall.
\newblock A new measure of rank correlation.
\newblock {\em Biometrika}, 30(1/2):81--93, 1938.

\bibitem{jin2024agentreview}
Yiqiao Jin, Qinlin Zhao, Yiyang Wang, Hao Chen, Kaijie Zhu, Yijia Xiao, and
  Jindong Wang.
\newblock Agentreview: Exploring peer review dynamics with llm agents.
\newblock {\em arXiv:2406.12708}, 2024.

\bibitem{KHAIRE20221060}
Utkarsh~Mahadeo Khaire and R.~Dhanalakshmi.
\newblock Stability of feature selection algorithm: A review.
\newblock {\em Journal of King Saud University - Computer and Information
  Sciences}, 34(4):1060--1073, 2022.

\bibitem{velivckovic2018graph}
Petar Veli{\v{c}}kovi{\'c}, Guillem Cucurull, Arantxa Casanova, Adriana Romero,
  Pietro Li{\`o}, and Yoshua Bengio.
\newblock Graph attention networks.
\newblock In {\em ICLR}, 2018.

\bibitem{jin2022code}
Yiqiao Jin, Yunsheng Bai, Yanqiao Zhu, Yizhou Sun, and Wei Wang.
\newblock Code recommendation for open source software developers.
\newblock In {\em Web Conference}, 2023.

\bibitem{kumar2019predicting}
Srijan Kumar, Xikun Zhang, and Jure Leskovec.
\newblock Predicting dynamic embedding trajectory in temporal interaction
  networks.
\newblock In {\em KDD}, pages 1269--1278, 2019.

\bibitem{jin2022towards}
Yiqiao Jin, Xiting Wang, Ruichao Yang, Yizhou Sun, Wei Wang, Hao Liao, and Xing
  Xie.
\newblock Towards fine-grained reasoning for fake news detection.
\newblock In {\em AAAI}, volume~36, pages 5746--5754, 2022.

\bibitem{yang2022reinforcement}
Ruichao Yang, Xiting Wang, Yiqiao Jin, Chaozhuo Li, Jianxun Lian, and Xing Xie.
\newblock Reinforcement subgraph reasoning for fake news detection.
\newblock In {\em KDD}, pages 2253--2262, 2022.

\bibitem{rozemberczki2021pytorch}
Benedek Rozemberczki, Paul Scherer, Yixuan He, George Panagopoulos, Alexander
  Riedel, Maria Astefanoaei, Oliver Kiss, Ferenc Beres, Guzm{\'a}n L{\'o}pez,
  Nicolas Collignon, et~al.
\newblock Pytorch geometric temporal: Spatiotemporal signal processing with
  neural machine learning models.
\newblock In {\em CIKM}, pages 4564--4573, 2021.

\bibitem{torgerson1952multidimensional}
Warren~S Torgerson.
\newblock Multidimensional scaling: I. theory and method.
\newblock {\em Psychometrika}, 17(4):401--419, 1952.

\bibitem{lee2022explaining}
Seongmin Lee, Sadia Afroz, Haekyu Park, Zijie~J Wang, Omar Shaikh, Vibhor
  Sehqal, Ankit Peshin, and Duen~Horng Chau.
\newblock Explaining website reliability by visualizing hyperlink connectivity.
\newblock In {\em 2022 IEEE Visualization and Visual Analytics (VIS)}, pages
  26--30. IEEE, 2022.

\bibitem{li2022visual}
Kevin Li, Haoyang Yang, Evan Montoya, Anish Upadhayay, Zhiyan Zhou, Jon
  Saad-Falcon, and Duen~Horng Chau.
\newblock Visual exploration of literature with argo scholar.
\newblock In {\em CIKM}, pages 4912--4916, 2022.

\bibitem{chomel2022polarization}
Victor Chomel, Nathana{\"e}l Cuvelle-Magar, Maziyar Panahi, and David
  Chavalarias.
\newblock Polarization identification on multiple timescale using
  representation learning on temporal graphs in eulerian description.
\newblock In {\em NeurIPS 2022 Temporal Graph Learning Workshop}, 2022.

\bibitem{shetty2005discovering}
Jitesh Shetty and Jafar Adibi.
\newblock Discovering important nodes through graph entropy the case of enron
  email database.
\newblock In {\em Proceedings of the 3rd international workshop on Link
  discovery}, pages 74--81, 2005.

\bibitem{seeger2003explaining}
Matthew~W Seeger and Robert~R Ulmer.
\newblock Explaining enron: Communication and responsible leadership.
\newblock {\em Management Communication Quarterly}, 17(1):58--84, 2003.

\bibitem{van2007essentials}
Cees~BM Van~Riel and Charles~J Fombrun.
\newblock {\em Essentials of corporate communication: Implementing practices
  for effective reputation management}.
\newblock Routledge, 2007.

\bibitem{yates1993control}
JoAnne Yates.
\newblock {\em Control through communication: The rise of system in American
  management}, volume~6.
\newblock JHU Press, 1993.

\bibitem{men2014strategic}
Linjuan~Rita Men.
\newblock Strategic internal communication: Transformational leadership,
  communication channels, and employee satisfaction.
\newblock {\em Management communication quarterly}, 28(2):264--284, 2014.

\bibitem{kovacs2019urban}
Zolt{\'a}n Kov{\'a}cs, Zsolt~Jen{\H{o}} Farkas, Tam{\'a}s Egedy, Attila~Csaba
  Kondor, Bal{\'a}zs Szab{\'o}, J{\'o}zsef Lennert, Dori{\'a}n Baka, and
  Bal{\'a}zs Koh{\'a}n.
\newblock Urban sprawl and land conversion in post-socialist cities: The case
  of metropolitan budapest.
\newblock {\em Cities}, 92:71--81, 2019.

\bibitem{skaf2022towards}
Wadie Skaf, Arzu Tosayeva, and D{\'a}niel~T V{\'a}rkonyi.
\newblock Towards automatic forecasting: Evaluation of time-series forecasting
  models for chickenpox cases estimation in hungary.
\newblock In {\em ISDA}, pages 1--10. Springer, 2022.

\end{thebibliography}

\newpage 

\newcounter{tempFigureCounter}
\setcounter{tempFigureCounter}{\value{figure}}


\beginsupplement

\section{Dataset Introduction}
\label{ref:dataset}
\vspace{1mm}
\paragraph{Datasets.} 
We used 9 publicly available datasets spanning 8 different domains to demonstrate \model's wide applicability across all of these subject areas. 
Table~\ref{tab:dataset} provides the statistics of the nine datasets. 

\begin{itemize}[leftmargin=.12in]
    \item \textbf{Reddit}
    ~\cite{jin2023predicting} encompasses YouTube videos shared across 29,461 subreddits over a five-year period, from January 2018 to December 2022. The dataset forms a bipartite graph with each node representing a video or a subreddit. Each edge in the graph indicates a video being shared in a subreddit, and its weight is determined by the frequency of sharing. 
    
    \item \textbf{Enron}
    ~\cite{klimt2004introducing} includes the email communication history of Enron Corporation from June 1999 to December 2001. Each node represents an employee and each edge represents an email between them. 
    
    \item \textbf{UN Comtrade}\footnote{\url{https://comtradeplus.un.org/}} (United Nations Comtrade database)~\cite{comtrade2010united} offers extensive global annual trade statistics. 
    Our analysis focuses on the annual export data from 1988 to 2022. Nodes represent countries and edges represent the logarithmic values of the annual export volumes between countries. 
    \item \textbf{HistWords-EN}. The \href{https://github.com/williamleif/histwords}{HistWords} embeddings is derived from the diachronic word embeddings trained using SGNS (Skip-Gram with Negative Sampling) on the Google N-Gram dataset~\cite{lin2012syntactic}, which uses English documents from the 1800s to the 1990s as the corpus. 
    Each node represents a word, and each edge represents word similarity. The detailed dataset construction process is described in Section~\ref{sec:HistWords}
    \item \textbf{HistWords-CN}~\cite{hamilton2016diachronic} is trained in the same manner as HistWords-EN using SGNS vectors of Chinese words from the Google N-Gram dataset over the period of 1950s to 1990s. 

    \item \textbf{Chickenpox}~\cite{rozemberczki2021chickenpox} features the weekly chickenpox cases in Hungary between January 2005 and January 2015. Nodes represent the counties, and edges are constructed based on geographical locations --- an edge exists between two counties if they are adjacent. The training objective is to predict the number of weekly cases in each county.

    \item \textbf{Ant}~\cite{mersch2013tracking} features ants behaviors over a 41 days' period. Nodes represent ants, and edges represent interactions between two ants.

    \item \textbf{DGraph}~\cite{huang2022dgraph} is a finance dataset about fraudster detection. Nodes represents \href{https://ir.finvgroup.com/}{Finvolution} users, which fall under 3 categories --- normal users, fraudsters, and background users (users who are not detection targets due to insufficient borrowing behaviors). An edge from one user to the other means that the user regards the other one as the emergency contact. We randomly sampled a subgraph with 100,000 nodes. 

    \item \textbf{Aging}~\cite{li2021improved} provides the human gene expression data at 37 ages spanning between 20 and 99 years. For each age, an aging-specific graph snapshot is constructed, in which nodes represent genes and edges represent interactions between genes. The edge weight represents the strength of the protein-protein interactions (PPIs) between two genes~\cite{newaz2020inference}. 
    
\end{itemize}




\begin{table}[!t]
\centering
\caption{Statistics of our datasets. ``Interval'' indicates the time interval for each snapshot. `\textbackslash{}' indicates that the snapshot interval is not constant.}
\label{tab:dataset}
\renewcommand{\arraystretch}{1.2}
\begin{tabular}{l|ccccc}
\toprule
\textbf{Datasets} & \textbf{Domains} &  \textbf{\#Nodes} & \textbf{\#Edges} & \textbf{\#Snapshots} & \textbf{Interval} \\ \midrule
\textbf{Reddit}~\cite{jin2023predicting} & Social Studies & 4,303,032 & 27,836,000 & 60 & 1 month \\
\textbf{DGraph}~\cite{huang2022dgraph} & Finance & 100,000 & 119,352 & 17 & 1 week \\
\textbf{HistWords-EN}~\cite{hamilton2016diachronic} & Linguistics & 100,000 & 14,539,140 & 20 & 10 years \\
\textbf{HistWords-CN}~\cite{hamilton2016diachronic} & Linguistics & 29,701& 763,100 & 5 & 10 years \\
\textbf{Aging}~\cite{li2021improved} & Genetics & 8,938 & 71,800 & 37 & \textbackslash{} \\
\textbf{Enron}~\cite{klimt2004introducing} & Communication Studies & 143 & 22,784 & 16 & 2 months \\
\textbf{Ant}~\cite{mersch2013tracking} & Ethology & 113 & 111,578 & 41 & 1 day \\
\textbf{UN Comtrade}~\cite{comtrade2010united} & International Relations & 107 & 162,322 & 35 & 1 year \\
\textbf{Chickenpox} & Epidemiology & 20 & 102 & 517 & 1 week \\
\bottomrule
\end{tabular}
\end{table}

\section{Method}
\label{sec:method}

\begin{figure*}[ht]
    \centering
    \includegraphics[width=0.95\textwidth]{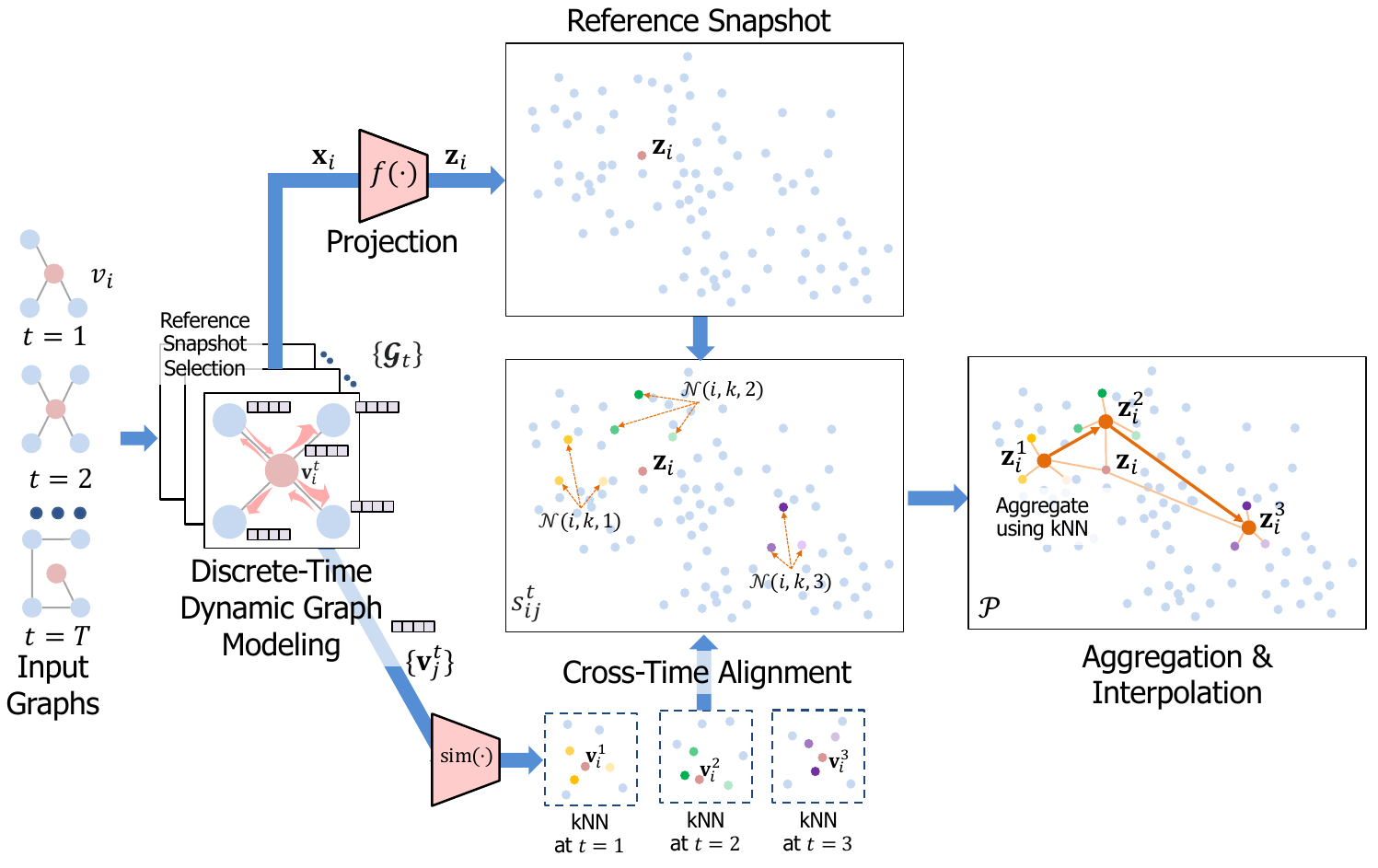} 
    \caption{Our proposed \model framework. }
    \label{fig:model}
\end{figure*}

In this section, we introduce our novel and computationally efficient framework \model for visualizing and analyzing dynamic graph embedding trajectories. 
Our framework effectively addresses the challenges associated with DGs mentioned in Section~\ref{sec:introduction}, including continuously evolving node embeddings and constant node addition and deletion. 
Figure~\ref{fig:model} and Algorithm~\ref{alg:algorithm} describe the workflow and the pseudocode of \model, respectively.


\vspace{1mm}
\subsection{Embedding Training}

Given the sequence of graph snapshots $\{G_t\}$, \model first learns a DTDG model using the joint training objective $\mathcal{L}$, 
which is the linear combination of the link prediction loss $\mathcal{L}_{\text{link}}^t$, node-level loss $\mathcal{L}_{\text{node}}^t$, and edge-level loss $\mathcal{L}_{\text{edge}}^t$. 
\begin{align}
\mathcal{L} &= \sum_{t \in [1, T]} \lambda_1 \mathcal{L}_{\text{link}}^t + \lambda_2 \mathcal{L}_{\text{node}}^t + \lambda_3 \mathcal{L}_{\text{edge}}^t. \label{eq:totalLoss}
\end{align}
Here, $\mathcal{L}_{\text{node}}^t$ (resp. $\mathcal{L}_{\text{edge}}^t$) can be defined as the mean squared error or cross-entropy loss between the predicted node (resp. edge) attributes and the ground-truth, depending on the problem formulation (\eg, linear regression or node/edge classification). $\lambda_1, \lambda_2, \lambda_3 \in \mathbb{R}$ denote hyperparameters that control the weights of each loss term. 
This process generates temporal node embeddings $\{\mathbf{V}^t\}_{t=1}^T$ across $T$ timestamps (Line \ref{line:Train}). 

\subsection{Embedding Visualization}
\begin{algorithm}[t]
\caption{DTDG embedding visualization. $\{\mathbf{V}^t\}$ is the set of temporal embedding matrix, where $\mathbf{V}^t \in \mathbb{R}^{|V^t| \times d}$. $\mathbf{X} \in \mathbb{R}^{|V'| \times d}$ is the static embedding matrix for the anchor nodes. $\operatorname{sim}(\cdot): \mathbb{R}^{d \times d} \rightarrow \mathbb{R}$ is a similarity measure. $\mathcal{N}(i, k, t)$ denotes the $k$ nearest neighbors of $v_i$ at time $t$ in the embedding space. $\operatorname{Agg}(\cdot)$ is an aggregation function. $\alpha$ is an interpolation factor.  $\mathbf{Z} = \{\mathbf{z}_i\}_{i=1}^{|V'|} \in \mathbb{R}^{|V'| \times p}$ is the $p$-dimensional projection of nodes $v_i \in V'$.}
\label{alg:algorithm}
\begin{algorithmic}[1]
\Require $\{\mathcal{G}^t\}$.
\Ensure Dynamic Graph Visualization $\mathcal{P}$.
\State{Train a DTDG model using objective $\mathcal{L}$ and derive $\{\mathbf{V}^t\}$} \Comment{Discrete-Time Dynamic Graph Model Training} \label{line:Train}
\State{Compute $\mathbf{X}$ for $v_i \in V'$} \label{line:RefEmb}
\State $\mathbf{Z} = f (\mathbf{X})$ \Comment{Compute $p$-dimensional projection of $V'$} \label{line:PorjFun}
\State Create $\mathcal{P}$ and project $\mathbf{Z} = \{\mathbf{z}_i\}_{i=1}^{|V'|}$ for $v_i \in V'$ onto $\mathcal{P}$ \label{line:Projection}
\For{$t \gets 1, \ldots, T$} \Comment{Cross-Time Alignment} \label{line:CrossTimeAlignment}
    \For{$v_i \in V^t$} \label{line:InnerLoopStart}
        \For{$v_j \in V' \setminus \{v_i\}$} 
            \State $s_{ij}^t \gets \operatorname{sim}(\mathbf{v}_i^t, \mathbf{v}_j^t)$ \Comment{Embedding Similarity} \label{line:CosSim}
            \EndFor \label{line:InnerLoopEnd}
        \State Compute $\mathcal{N}(i, k, t)$ according to $\{s_{ij}^t\}$ \label{line:knn}
        \State $\mathbf{\hat{z}}_i^t \gets \operatorname{Agg}(\{\mathbf{z}_j | v_j \in \mathcal{N}(i, k, t)\})$ \label{line:Aggregation} \Comment{Aggregation}
        \State $\mathbf{z}_i^t =
        \begin{cases}
            \alpha \cdot \mathbf{z}_i + (1 - \alpha) \cdot \mathbf{\hat{z}}_i^t & \text{if\ } v_i \in V' \\
            \mathbf{\hat{z}}_i^t & \text{otherwise}
        \end{cases}$ \label{line:Interpolation} \Comment{Interpolation}
        \State Project $\mathbf{z}_i^t$ onto $\mathcal{P}$.
    \EndFor
\EndFor 
\end{algorithmic}
\end{algorithm}
 
A major challenge in embedding trajectories visualization is cross-time alignment, as the DTDG embeddings from different snapshots reside in distinct embedding spaces and are not directly comparable with each other~\cite{szymanski2017temporal}. To address this challenge, we construct a uniform reference frame for the embedding projection of all snapshots using carefully selected \emph{anchor nodes} $V'$. The \emph{anchor nodes} are selected from the set of nodes present in $V^0$ to ensure meaningful cosine similarity computation in each snapshot. 
$\mathbf{X}$, the node embeddings of $V'$, can be derived from a subset of any temporal embedding $\mathbf{V}^t$ trained on time $t$ (Line~\ref{line:RefEmb}). \model is based on the assumption that the embeddings of nodes in $V'$ do not undergo significant changes over time~\cite{dadu2023application}. 

We then employ the projection function $f(\cdot)$ to derive the $p$-dimensional representations $\mathbf{Z}$ (Line~\ref{line:PorjFun}). 
The choice of $f(\cdot)$ provides flexibility, allowing various projection algorithms that preserve the node-node proximity in the embedding space such as Principal Component Analysis (PCA)~\cite{hotelling1933analysis}, t-SNE~\cite{van2008visualizing}, H-SNE~\cite{pezzotti2016hierarchical}, UMAP~\cite{mcinnes2018umap}, locally linear embedding (LLE)~\cite{roweis2000nonlinear}, and Isomap~\cite{tenenbaum2000global} to be employed. 
This initial projection serves as a steady topological foundation that ensures consistency across all timestamps. 
As \model\ progresses through each timestamp $t$, it updates the visual representations of each node $v_i\in V^t$ in $G^t$, considering its new positions (Lines \ref{line:InnerLoopStart}-\ref{line:InnerLoopEnd}).
To this end, we identify $v_i$'s $k$ nearest anchor nodes $v_j\in V'$ based on the similarity between the temporal embeddings of $v_i$ and $v_j$ (Lines \ref{line:CosSim}, \ref{line:knn}).
We then aggregate the visual representations of the neighboring anchor nodes $v_j$ to determine the new position for $v_i$ (Line~\ref{line:Aggregation}).
This method efficiently aligns nodes across different timestamps and allows for the inference of new nodes by aggregating information from anchor nodes. 
Therefore, \model\ can seamlessly incorporate new nodes into the visualization space, such as newly formed COVID-related online communities on social platforms during the COVID-19 pandemic, or the inclusion of new words into a vocabulary in diachronic linguistic analysis. 
To ensure coherence and smooth transitions between timestamps, the final node projection is obtained by interpolation, combining the aggregated projection $\mathbf{\hat{z}}_i^t$ with $v_i$'s static embedding $\mathbf{z}_i$ (Line~\ref{line:Interpolation}).





\vspace{1mm}
\subsection{Analytics Module.}
\label{sec:analytics}
We employ micro-level and macro-level measures to quantify the structural shifts in both \emph{local} and \emph{global} topology. 

\paragraph{Measuring Micro-level Changes.}
To quantify the micro-level changes in the \emph{local} topology of each node, we employ two similarity measures: Jaccard index ($\operatorname{Jaccard}_n$)~\cite{jaccard1912distribution} and Rank-biased Overlap (RBO)~\cite{webber2010similarity,oh2022rank}. 
The Jaccard index quantifies the agreement between the closest $n$ nodes of a given node $i$ at time $(t-1)$ and those at time $t$ in the embedding space. It is calculated as the intersection size between two sets divided by the size of their union.
\begin{equation}
  \operatorname{Jaccard}_n(i, t)=\frac{\mathcal{N}(i, n, t-1) \cap \mathcal{N}(i, n, t)}{\mathcal{N}(i, n, t-1) \cup \mathcal{N}(i, n, t)},
\end{equation}
where $\mathcal{N}(i, n, t)$ indicates the closest $n$ nodes of $v_i$ sorted in ascending order based on their distance from $v_i$ at time $t$. 
The resulting $\mathrm{Jaccard}_n$ ranges from 0 to 1 and is agnostic to the ordering of the top-$n$ nodes. A $\mathrm{Jaccard}_n$ close to 1 during the period $[t-1, t]$ indicates minimal changes in the node's \emph{local} topology in the embedding space. 

As a complementary measure, Ranked Bias Overlap (RBO) considers the absolute ranking of nodes. 
RBO gradually incorporates lower-ranked nodes while also accounting for the top-ranked ones. 
\begin{equation}
  \operatorname{RBO}(i, m, t)=(1-p) \sum_{d=1}^{m} p^{d-1} \frac{|\mathcal{N}(i, m, t-1) \cap \mathcal{N}(i, m, t)|}{d},
\end{equation}
where $m$ represents the maximum depth of the ranked list considered, and $p \in [0, 1]$ is the damping factor that determines the weight assigned to the top of the list.
A higher value of $p$ (closer to 1) assigns more significance to the top of the list. In our experiments, we set $p$ to 0.9. 
The RBO metric ranges from 0 to 1, with a higher value indicating greater similarity in the node ordering between the two lists.
Intuitively, if a node's RBO is close to 1 during the period $(t-1)$ to $t$, the node's \emph{global} topology in the input DG has undergone minimal changes.

\begin{figure*}[ht]
    \centering
    \includegraphics[width=0.95\textwidth]{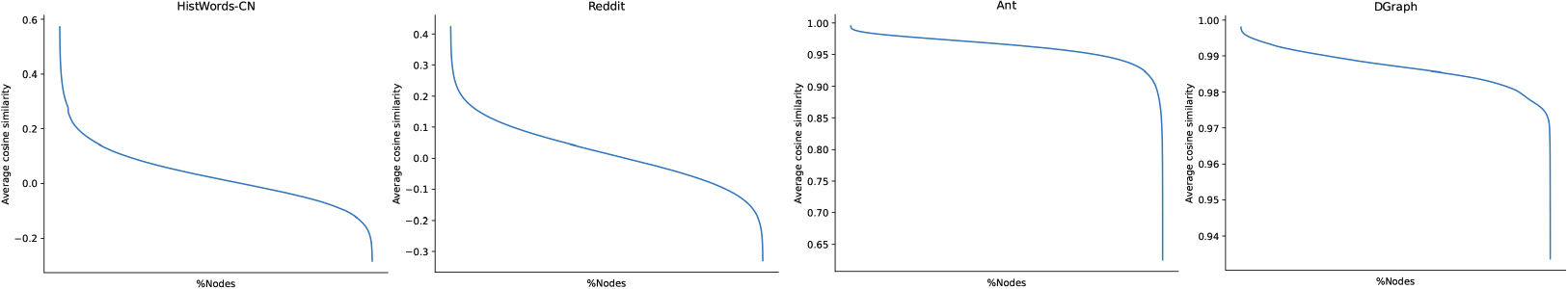}
    \caption{Distribution of average cosine similarity of the four datasets: HistWords-CN, Reddit, Ant, and DGraph. The x- and y- axis represents the percentage of the nodes in the dataset and cosine similarities, respectively.}
    \label{fig:CosSimDist}
\end{figure*}

It is worth noting that alternative ranking evaluation measures, such as Spearman's rank correlation coefficient~\cite{spearman1904proof} and Kendall's tau~\cite{kendall1938new,jin2024agentreview}, exist. However, these measures do not explicitly differentiate the importance of the ranks at different positions in the list and are sensitive to small perturbations of rankings, particularly towards the middle of the list~\cite{KHAIRE20221060}. To demonstrate this, Supp. Table~\ref{fig:CosSimDist} shows the distribution of average cosine similarity for all nodes in the four datasets, HistWords-CN, Reddit, Ant, and DGraph. We observe that the cosine similarity usually plateaus in the middle range, suggesting a large number of nodes with highly similar cosine similarity. 

Consequently, they cannot accurately reflect the extent to which the local neighbors of a node have changed. 
Moreover, these are also mainly focused on conjoint rankings~\cite{webber2010similarity} where both lists consist of the same set of items, making them less suitable for scenarios where the set of nodes in adjacent snapshots are different due to new nodes constantly being added for comparison. 
In contrast, RBO and Jaccard index are more responsive to changes in the top portion of two ranked lists and can be applied to indefinite ranking scenarios, which aligns well with our objectives, as we emphasize the importance of top-$n$ nodes for assessing changes of the local neighbors of each node in the visualization.


\vspace{1mm}
\paragraph{Measuring Macro-level Changes.} 
To assess the changes in global topology, we introduce a novel metric called \emph{Normalized Average Rank Change (NARC)}, which builds upon the Average Rank Change (ARC) metric:
\begin{align}
    &\operatorname{ARC}(i, t) = \frac{1}{N^{t}} \sum_{j=1}^{N^{t}} |r_{ij}^{t} - r_{ij}^{t-1}|, 
    \label{eq:arc} \\
    &\operatorname{NARC} = \frac{1}{T} \sum_{t=1}^{T} \frac{1}{N^t - 1} \sum_{i=1}^{N^t} \operatorname{ARC}(i, t),
    \label{eq:narc}
\end{align}
where $N^{t} = |V^{t} \cap V^{t-1}|$ represents the number of nodes jointly present in both time $(t-1)$ and $t$. $\operatorname{ARC}(i, t)$ measures the changes of a node $i$'s nearest neighbors in the period $[t-1, t]$, where a greater $\operatorname{ARC}(i, t)$ indicates a larger change in $i$'s topology. The NARC metric is an aggregated metric across all nodes and timestamps. 
By normalizing each $\operatorname{ARC}(i, t)$ by a factor of $N^t - 1$, we make the NARC metric comparable across datasets with different sizes. The NARC metric provides a comprehensive assessment of the changes in the \emph{global} topology across all nodes and timestamps, offering valuable insights into the dynamic nature of the evolving network.

To measure the absolute movements of node embeddings in the embedding space over time, we use the L1 and L2 distances between the embeddings of each node $v_i$ in adjacent timestamps:
\begin{align}
  \operatorname{L}_p &= \frac{1}{T-1} \frac{1}{N^t} \sum_{t=1}^{T-1} \sum_{i=1}^{N^t} \| \mathbf{h}_i^t -  \mathbf{h}_i^{t-1} \|_p, \quad p \in [1, 2], \label{eq:Movement}
\end{align}
where $\mathbf{h}_i^t$ is an embedding at time $t$. Here, we consider $\mathbf{h}_i^t$ being one of $\mathbf{v}_i^t, \mathbf{\tilde{v}}_i^{t-1}$, and $\mathbf{z}_i^t$, where $\mathbf{v}_i^t$ is the original embedding, $\mathbf{\tilde{v}}_i^{t-1} = \mathbf{v}_i^t / \|\mathbf{v}_i^t\|$ is the normalized embedding, and $\mathbf{z}_i^t$ is the projected embeddings. 

Finally, we extend the RBO metric to a macro-level version, which is called macro-level RBO, as follows:
\begin{equation}
  \operatorname{RBO}_{\text{macro}}(i, m, t)= \frac{1}{T} \sum_{t=1}^{T} \frac{1}{N^t} \sum_{i=1}^{N^t} \operatorname{RBO}(i, m, t).
\end{equation}

\section{Related Works}
\label{sec:related}

\subsection{Graph Neural Network}
Graph neural networks~\cite{velivckovic2018graph} have emerged as a powerful framework for modeling complex relationships in graph-structured data. In particular, dynamic graph models, which capture temporal dynamics in evolving systems, have been successfully applied in analyzing various domains such as communication networks~\cite{sankar2020dysat}, transaction networks~\cite{jin2022code}, social networks~\cite{kumar2019predicting,jin2023predicting,jin2022towards,yang2022reinforcement}, disease control~\cite{rozemberczki2021pytorch}, and international trade~\cite{monken2021graph}.

\subsection{Visualization}

Visualization is a popular approach for model analytics due to its user-friendly and intuitive nature, which allows researchers and analysts to easily comprehend complex temporal relationships.  Techniques such as Principal Component Analysis (PCA)~\cite{hotelling1933analysis}, t-Distributed Stochastic Neighbor Embedding (t-SNE)~\cite{van2008visualizing}, Multidimensional Scaling (MDS)~\cite{torgerson1952multidimensional}, and Uniform Manifold Approximation and Projection (UMAP)~\cite{mcinnes2018umap} have been widely used to represent high-dimensional data in a lower-dimensional space by preserve the structural relationships of the original data. 
Despite these advancements, there is still a need for visualization techniques that can effectively capture and represent the dynamics of evolving graph data over an extended period of time. 
Although researchers have explored visualization techniques for graphs, existing works usually focus on static graphs~\cite{lee2022explaining, li2022visual} or consecutive graph snapshots~\cite{chomel2022polarization}, limiting their ability to showcase the trajectory of node embeddings over time~\cite{chomel2022polarization}. This limitation hinders the comprehensive understanding of how nodes evolve and interact within the graph structure. 

\begin{figure*}[ht]
    \centering
    \includegraphics[width=0.95\textwidth]{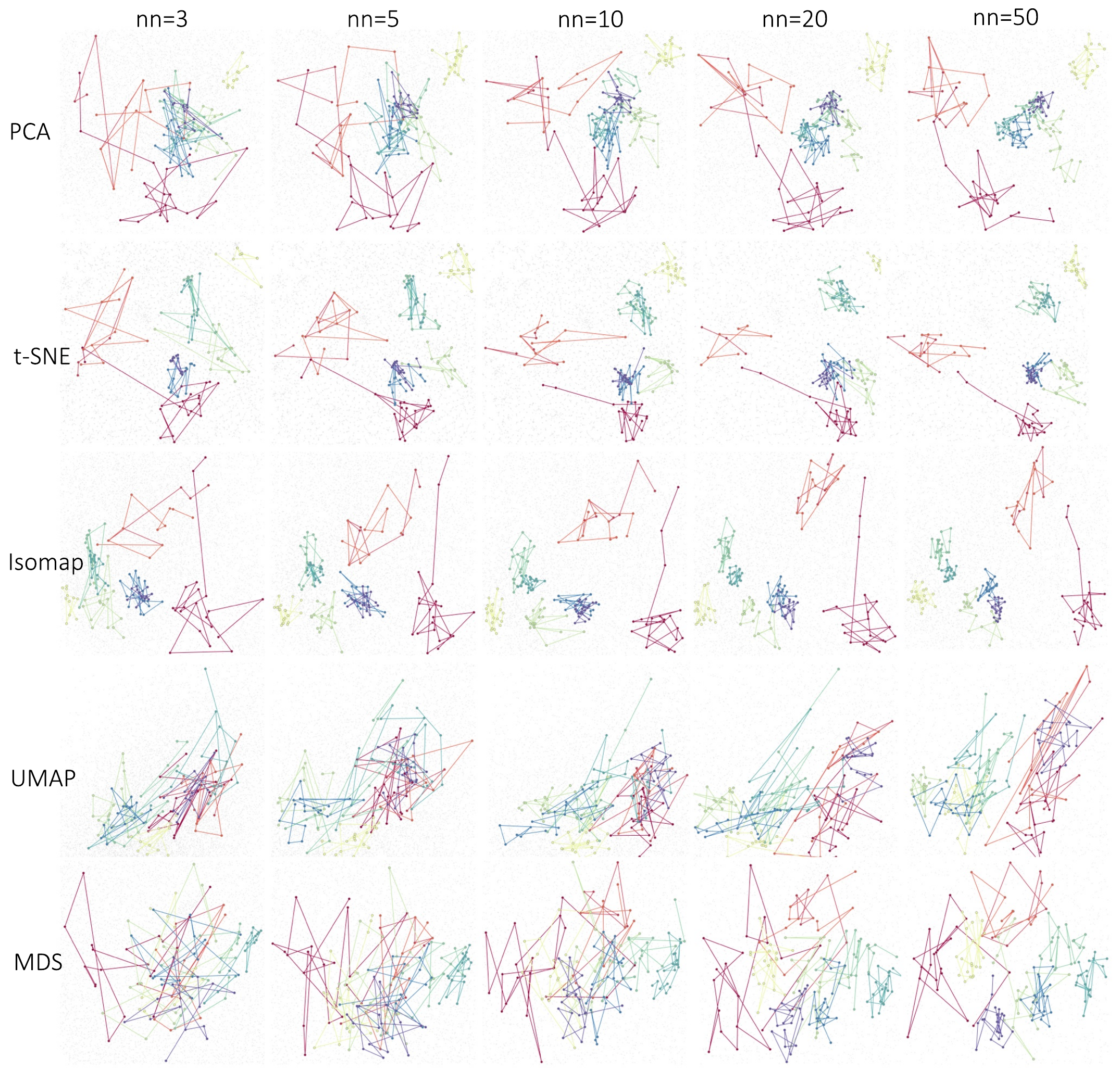} 
    \caption{Comparison of varying $f(\cdot)$ (Line~\ref{line:Projection}) and $k$ (Line~\ref{line:knn}) in Algorithm~\ref{alg:algorithm}. 
    }
    \label{fig:Sensitivity}
    \vspace{-0.5cm}
\end{figure*}

\begin{table}
\small
\centering
\vspace{-0.1cm}
\caption{Notations used in this paper}
\vspace{-0.2cm}
\label{tab:notations}
\renewcommand{\arraystretch}{1.2}
\begin{tabularx}{0.49\textwidth}{l|X} 
\toprule
\textbf{Notation} & \textbf{Description} \\ 
\midrule
$G$ & A static graph \\
$G^t$ & A graph snapshot at timestamp $t$ \\
$V, E$ & Sets of nodes and edges \\
$V^t, E^t$ & Sets of nodes and edges in each snapshot $G^t$\\
$V'$ & Set of anchor nodes \\
$v_i$ & A node $i$ \\
\multirow{2}{*}{$\mathcal{N}(i, k, t)$} & $v_i$'s list of $k$ nearest neighbors in $\mathbf{V}^t$, sorted in descending order \\
$\mathbf{v}_i^t$ & Temporal node embedding of $v_i$ at timestamp $t$ \\
$\mathbf{V}^t$ & Temporal node embeddings for $V^t$ at timestamp $t$ \\ 
$\mathbf{X}$ & Anchor embeddings \\
\midrule
$\mathcal{L}$ & Training Objective \\
$\alpha, \lambda_1, \lambda_2$ & Hyperparameters \\
\bottomrule
\end{tabularx}
\vspace{-0.2cm}
\end{table}

\begin{table*}[htbp]
\caption{Evolution of word associations from the 1950s to the 1990s. The term ``gay'' initially carried connotations with happiness and fortune, but underwent a decline in positivity during the 1970s as its association with homosexuality became more prevalent.
}
\label{tab:gaywords}
\small
\begin{tabular}{l|p{2.5cm}|p{2.5cm}|p{2.5cm}|p{2.5cm}|p{2.5cm}}
\toprule
Word & 1950 & 1960 & 1970 & 1980 & 1990 \\
\midrule
\multirow{4}{*}{happy} & happier, fortunate, glad, lucky, delighted & happier, pleasant, lucky, loved, delighted & glad, happier, fortunate, longed, delighted & glad, delighted, happier, pleasant, fortunate & glad, happier, delighted, eager, lucky \\
\midrule
\multirow{4}{*}{delighted} & glad, surprised, astonished, pleased, gratified & gratified, surprised, astonished, glad, amused & glad, surprised, astonished, gratified, pleased & glad, surprised, happy, amused, astonished & surprised, glad, pleased, happy, astonished \\
\midrule
\multirow{3}{*}{gay} & charming, lovely, beautiful, elegant, bright & elegant, charming, cheerful, lovely, witty & charming, cheerful, clubs, forlorn, ugly & boys, clubs, lovers, charming, men & men, victims, violence lesbian, bisexual\\
\midrule
\multirow{3}{*}{homosexual} & sex, intimacy, prostitution, males, females & sex, cruelties, immoral, notorious, scandalous & sex, unmarried, gender, adultery, immoral & women, unmarried, gay, immoral, sex & gay, males, immoral, illegal, sex \\
\midrule
\multirow{3}{*}{lesbian} & \textbackslash{} & \textbackslash{} & vehemently, clubs, gang, gay, dance & feminist, gay, sexuality, identities, women & gay, women, female, victims, violence \\
\bottomrule
\end{tabular}
\end{table*}

\begin{table}
\caption{Average $\operatorname{Jaccard}_{100}$ of subreddits in each year. }
\label{tab:reddit}
\centering
\begin{tabular}{p{5cm}|ccccc}
\toprule
Subreddit & 2018 & 2019 & 2020 & 2021 & 2022 \\
\midrule
videos & \ApplyGradient{0.118} & \ApplyGradient{0.211} & \ApplyGradient{0.206} & \ApplyGradient{0.258} & \ApplyGradient{0.256} \\
YouTubeSubscribeBoost & \ApplyGradient{0.244} & \ApplyGradient{0.389} & \ApplyGradient{0.296} & \ApplyGradient{0.295} & \ApplyGradient{0.316} \\
YouTube\_startups & \ApplyGradient{0.233} & \ApplyGradient{0.339} & \ApplyGradient{0.289} & \ApplyGradient{0.341} & \ApplyGradient{0.312} \\
AdvertiseYourVideos & \ApplyGradient{0.287} & \ApplyGradient{0.349} & \ApplyGradient{0.292} & \ApplyGradient{0.313} & \ApplyGradient{0.332} \\
SmallYoutubers & \ApplyGradient{0.267} & \ApplyGradient{0.356} & \ApplyGradient{0.304} & \ApplyGradient{0.306} & \ApplyGradient{0.354} \\
GetMoreViewsYT & \ApplyGradient{0.284} & \ApplyGradient{0.311} & \ApplyGradient{0.269} & \ApplyGradient{0.280} & \ApplyGradient{0.305} \\
\midrule
gaming & \ApplyGradient{0.263} & \ApplyGradient{0.280} & \ApplyGradient{0.285} & \ApplyGradient{0.272} & \ApplyGradient{0.249} \\
videogames & \ApplyGradient{0.303} & \ApplyGradient{0.275} & \ApplyGradient{0.324} & \ApplyGradient{0.322} & \ApplyGradient{0.292} \\
pcgaming & \ApplyGradient{0.258} & \ApplyGradient{0.236} & \ApplyGradient{0.277} & \ApplyGradient{0.266} & \ApplyGradient{0.223} \\
YouTubeGamers & \ApplyGradient{0.279} & \ApplyGradient{0.364} & \ApplyGradient{0.291} & \ApplyGradient{0.343} & \ApplyGradient{0.397} \\
PromoteGamingVideos & \ApplyGradient{0.317} & \ApplyGradient{0.423} & \ApplyGradient{0.336} & \ApplyGradient{0.383} & \ApplyGradient{0.407} \\
gamingvids & \ApplyGradient{0.318} & \ApplyGradient{0.329} & \ApplyGradient{0.295} & \ApplyGradient{0.296} & \ApplyGradient{0.300} \\
GlobalOffensive & \ApplyGradient{0.071} & \ApplyGradient{0.111} & \ApplyGradient{0.162} & \ApplyGradient{0.093} & \ApplyGradient{0.080} \\
apexlegends & N/A & \ApplyGradient{0.114} & \ApplyGradient{0.110} & \ApplyGradient{0.099} & \ApplyGradient{0.073} \\
leagueoflegends & \ApplyGradient{0.069} & \ApplyGradient{0.044} & \ApplyGradient{0.057} & \ApplyGradient{0.062} & \ApplyGradient{0.038} \\
Minecraft & \ApplyGradient{0.040} & \ApplyGradient{0.115} & \ApplyGradient{0.136} & \ApplyGradient{0.143} & \ApplyGradient{0.072} \\
\bottomrule
\end{tabular} \\
\begin{tabular}{p{5cm}|ccccc}
\toprule
Subreddit & 2018 & 2019 & 2020 & 2021 & 2022 \\
\hline
kpop & \ApplyGradient{0.066} & \ApplyGradient{0.082} & \ApplyGradient{0.129} & \ApplyGradient{0.164} & \ApplyGradient{0.151} \\
popheads & \ApplyGradient{0.170} & \ApplyGradient{0.160} & \ApplyGradient{0.181} & \ApplyGradient{0.221} & \ApplyGradient{0.195} \\
indieheads & \ApplyGradient{0.221} & \ApplyGradient{0.258} & \ApplyGradient{0.263} & \ApplyGradient{0.278} & \ApplyGradient{0.274} \\
Music & \ApplyGradient{0.187} & \ApplyGradient{0.224} & \ApplyGradient{0.291} & \ApplyGradient{0.339} & \ApplyGradient{0.331} \\
hiphopheads & \ApplyGradient{0.232} & \ApplyGradient{0.296} & \ApplyGradient{0.308} & \ApplyGradient{0.333} & \ApplyGradient{0.276} \\
listentothis & \ApplyGradient{0.234} & \ApplyGradient{0.325} & \ApplyGradient{0.359} & \ApplyGradient{0.374} & \ApplyGradient{0.338} \\
hiphop & \ApplyGradient{0.266} & \ApplyGradient{0.363} & \ApplyGradient{0.424} & \ApplyGradient{0.442} & \ApplyGradient{0.312} \\
rap & \ApplyGradient{0.282} & \ApplyGradient{0.413} & \ApplyGradient{0.421} & \ApplyGradient{0.417} & \ApplyGradient{0.296} \\
\midrule
sports & \ApplyGradient{0.090} & \ApplyGradient{0.111} & \ApplyGradient{0.101} & \ApplyGradient{0.101} & \ApplyGradient{0.084} \\
nba & \ApplyGradient{0.121} & \ApplyGradient{0.109} & \ApplyGradient{0.089} & \ApplyGradient{0.082} & \ApplyGradient{0.076} \\
nfl & \ApplyGradient{0.077} & \ApplyGradient{0.112} & \ApplyGradient{0.061} & \ApplyGradient{0.049} & \ApplyGradient{0.050} \\
MMA & \ApplyGradient{0.103} & \ApplyGradient{0.086} & \ApplyGradient{0.090} & \ApplyGradient{0.088} & \ApplyGradient{0.070} \\
SquaredCircle & \ApplyGradient{0.065} & \ApplyGradient{0.051} & \ApplyGradient{0.058} & \ApplyGradient{0.085} & \ApplyGradient{0.086} \\
\midrule
politics & \ApplyGradient{0.384} & \ApplyGradient{0.378} & \ApplyGradient{0.423} & \ApplyGradient{0.313} & \ApplyGradient{0.290} \\
The\_Donald & \ApplyGradient{0.331} & \ApplyGradient{0.375} & \ApplyGradient{0.188} & N/A & N/A \\
WayOfTheBern & \ApplyGradient{0.385} & \ApplyGradient{0.405} & \ApplyGradient{0.447} & \ApplyGradient{0.441} & \ApplyGradient{0.308} \\
\bottomrule
\end{tabular}
\end{table}

\section{Additional Experimental Results}

\subsection{Enron: Email Communication}
\label{sec:Enron}
\begin{figure*}[htbp]
    \centering
    \includegraphics[width=0.95\textwidth]{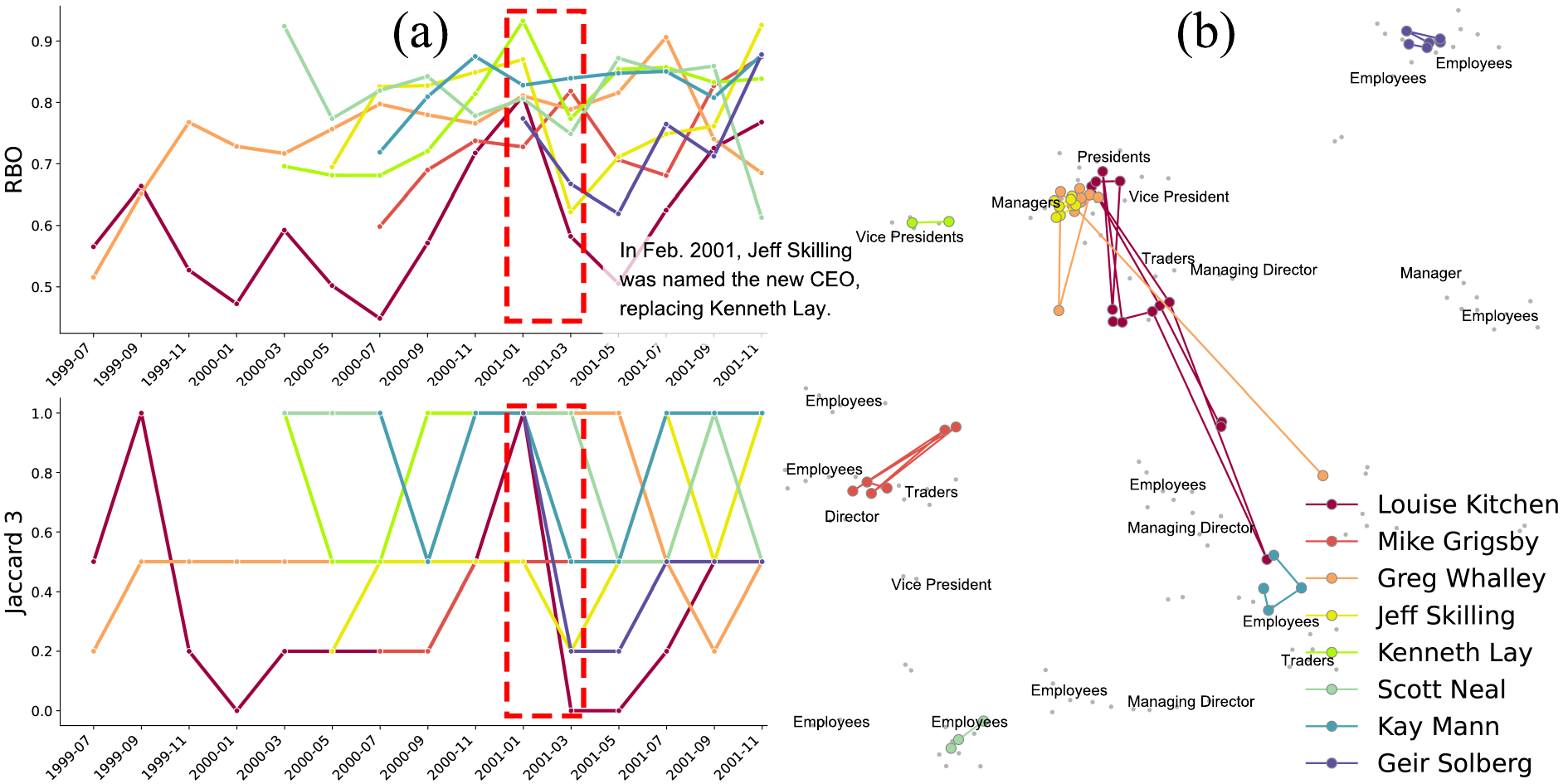} 
    \caption{Visualization of the Enron dataset~\cite{klimt2004introducing}. (a) RBO and $\operatorname{Jaccard}_3$ during the projection time period. (b) Dynamic visualization of the node trajectories. The positions of employees are annotated. 
    Employees occupying managerial/president positions (Mike Grigsby, Louise Kitchen and Greg Whalley) other than the CEOs interact with employees at various levels, resulting in more diverse trajectories compared to CEOs (Kenneth Lay and Jeffrey Skilling) and regular employees (\eg Scott Neal). 
    Furthermore, we observe a decline in both RBO and $\operatorname{Jaccard}_3$ from January to March 2001, suggesting a significant shift in communication partners among all employees. This decline aligns with the CEO transition in February 2001. }
    \label{fig:enron}
    \vspace{-0.5cm}
\end{figure*}
Enron Corporation, founded by Kenneth Lay in 1985, was a prominent energy company until its notorious collapse due to an institutionalized and systematic accounting fraud~\cite{klimt2004introducing, shetty2005discovering}. 
In February 2001, Jeffrey Skilling became Enron's CEO, initiating a period characterized by aggressive and intricate accounting practices~\cite{seeger2003explaining}. 
Jeffrey resigned in December 2001, shortly before Enron's downfall. Studies show that Enron's collapse can be attributed to a failure of responsible communication~\cite{seeger2003explaining}. Lay and Skilling were only partially aware of the financial misconduct of their subordinates. 
In this study, we investigate the email communication network from June 1999 to December 2001 to shed light on the internal communication patterns that contribute to Enron's failure.
Our visualization provides valuable insights into these communication patterns, particularly highlighting the trajectories of CEOs like Kenneth Lay and Jeffrey Skilling. 

In Fig.~\ref{fig:enron}b, the trajectories of Lay and Skilling indicate relatively static communication communities, confined within a small range that mainly involves a limited number of vice presidents and managers. These patterns reflect their shortcomings in two-way communication, as demonstrated in previous studies~\cite{shetty2005discovering, seeger2003explaining} --- the failure to deliver honest, ethical messages to employees and the lack of awareness regarding company operations. 
Meanwhile, ordinary employees such as Geir Solberg, Kay Mann, and Scott Neal have specific job functions that confine their communication to their respective teams, resulting in relatively fixed communication patterns with a limited number of partners and trajectories with less variability.

In contrast, managers and presidents play crucial roles in facilitating communication between upper management and ordinary employees~\cite{van2007essentials, yates1993control, men2014strategic}, resulting in more diverse interactions with individuals at different levels within the organization. 
Previous analyses identified the top three influential nodes in the Enron dataset as Louis Kitchen (President), Mike Grigsby (Manager), and Greg Whalley (President) according to graph entropy~\cite{shetty2005discovering}. Their trajectories in the visualization exhibit broader and more diverse patterns, involving different individuals in different positions over different time periods. This reflects their extensive responsibilities and significant roles in managing the organization. 

Moreover, as shown in Fig.~\ref{fig:enron}a, the decline in RBO and $\operatorname{Jaccard}_3$ across all employees during the CEO transition from Lay to Skilling (January - March 2001) highlights the impact of leadership changes on communication dynamics within the organization. 
\model uncovers differences in communication patterns among employees, offering a novel perspective on organizational structure and dynamics. Understanding these patterns provides valuable insights for future research and organizational management.


\subsection{Dynamic Temporal and Spatial Modeling of Chickenpox Spread in Hungary}
\begin{figure*}[ht]
    \centering
    \includegraphics[width=0.75\textwidth]{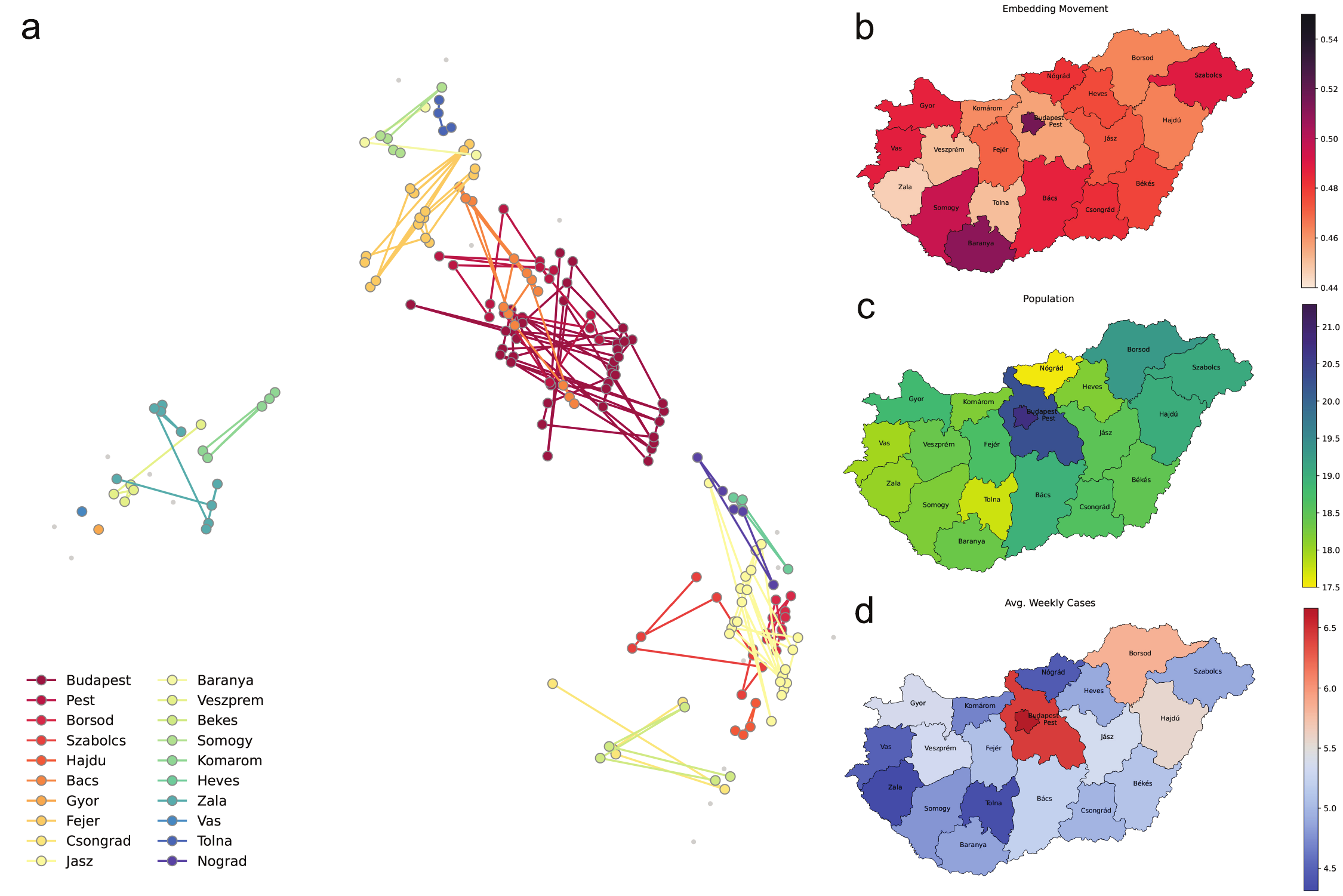} 
    \caption{\textbf{a.} Trajectories of the 19 counties and the capital Budapest in the Chickenpox dataset~\cite{rozemberczki2021chickenpox}. The most (least) populous counties are plotted using colors to the red (purple) side of the spectrum, and vice versa. \textbf{b.} Embedding movements of the 20 nodes. \textbf{c/d.} Population and average weekly chickenpox cases of all counties.}
    \label{fig:Chickenpox}
    \vspace{-0.5cm}
\end{figure*}

In epidemiology forecasting, dynamic graph models can potentially enhance our understanding and prediction of disease spread. By incorporating temporal and spatial dynamics, these models can capture the intricate interplay between population density, geography, and mobility patterns, all of which play critical roles in disease transmission. However, the successful application of DTDG models in epidemiology forecasting relies not only on the accuracy and robustness of the models, but also on our ability to interpret and understand their mechanisms of operation. In this regard, the use of visualization techniques becomes crucial. 

In this study, we use the Hungary Chickenpox dataset~\cite{rozemberczki2021chickenpox}, which includes the weekly chickenpox cases in Hungarian counties and the capital Budapest between 2005 and 2015. From the trajectories in Fig.~\ref{fig:Chickenpox}a, we found that the capital city Budapest stands out as the node with the most movements due to its high population and . 
As the second most populous county in the country, Pest exhibits a trajectory that significantly overlaps with Budapest at each snapshot, indicating a high seasonality in the number of cases. 
This overlap can be attributed to factors such as geographical locations and suburbanization in the metropolitan area of Budapest, which caused considerable population movements between the two regions. 
According to a census in 2011~\cite{kovacs2019urban}, nearly 60\% of commuters living in the suburban zone of Pest work in Budapest. Such population overlap can facilitate the spread of diseases. B\'{a}cs-Kiskun, a bordering state of Pest, moves towards Pest in winter, especially the middle of December, but away from it in summer, indicating a periodicity between the winter surge and the summer decline~\cite{rozemberczki2021chickenpox, skaf2022towards}. 
In contrast, counties such as Tolna, Vas, Zala, and Heves, which are among the five least populated counties, form their own clusters with movements limited to the lower left of the plot, indicating a lower susceptibility to diseases due to smaller populations and fewer demographic movements. 

\renewcommand{\figurename}{Figure}
\renewcommand{\tablename}{Table}

\end{document}